\pdfoutput=1
\documentclass[11pt]{article}

\usepackage[final]{acl}

\usepackage{times}
\usepackage{latexsym}
\usepackage{subfig}
\usepackage{graphicx}
\usepackage{subcaption}
\usepackage{amsmath}
\usepackage{multirow}
\usepackage{tabularx}
\usepackage{enumitem}
\usepackage{pbox}
\usepackage{float}
\usepackage{hyperref}
\usepackage{setspace}
\usepackage{diagbox}
\usepackage{float}
\usepackage{subcaption}
\usepackage{caption}
\usepackage{booktabs}
\usepackage{bbding}
\usepackage{mathtools}
\usepackage{color, colortbl}
\definecolor{grey}{rgb}{0.898,0.898,0.898}
\usepackage{amssymb}% http://ctan.org/pkg/amssymb
\usepackage{pifont}% http://ctan.org/pkg/pifont
\newcommand{\proposed}{\textsc{Simplot}}
\newcommand{\prompt}{\textit{Human-oriented chart instruction}}
\newcommand{\equal}{\thanks{These authors contributed equally.}}
\usepackage[T1]{fontenc}
\usepackage[utf8]{inputenc}

\usepackage{microtype}

\usepackage{inconsolata}

\title{\textsc{Simplot}: Enhancing Chart Question Answering by Distilling Essentials}

\author{
\textbf{Wonjoong Kim}\equal, \ 
\textbf{Sangwu Park}\footnotemark[1], \ 
\textbf{Yeonjun In}, \ 
\textbf{Seokwon Han}, \ 
\textbf{Chanyoung Park}\thanks{Corresponding author.}\\
KAIST \qquad \\
\texttt{\{wjkim, sangwu.park, yeonjun.in, dulwich10, cy.park\}@kaist.ac.kr}
}

\begin{document}
\maketitle
\begin{abstract}

Recently, interpreting complex charts with logical reasoning has emerged as challenges due to the development of vision-language models. A prior state-of-the-art (SOTA) model has presented an end-to-end method that leverages the vision-language model to convert charts into table format utilizing Large Language Model (LLM) for reasoning. However, unlike natural images, charts contain a mix of essential and irrelevant information required for chart reasoning, and we discover that this characteristic can lower the performance of chart-to-table extraction. In this paper, we introduce \proposed, a method designed to extract only the elements necessary for chart reasoning. The proposed method involves two steps: 1) training to mimic a simple plot that contains only the essential information from a complex chart for table extraction, followed by 2) performing reasoning based on the table. Our model enables accurate chart reasoning without the need for additional annotations or datasets, and its effectiveness is demonstrated through various experiments\footnote{Our source code is available at \url{https://github.com/sangwu99/Simplot}.}. 

\end{abstract}

\section{Introduction}
\label{sec:introduction}

The rapid advancements in vision-language models have accelerated a wave of research into models capable of handling data that integrates both images and text \citep{zhang2021vinvl, zhou2020unified, kim2023llm4sgg, kim2024adaptive}, thus undertaking a variety of tasks. Among these tasks, the interest in models capable of advanced reasoning has significantly increased. This increasing field has seen considerable success in addressing visual question answering (VQA) tasks targeted at natural images, marking a trend towards models that can engage in complex reasoning based on images \citep{antol2015vqa, shao2023prompting, garderes2020conceptbert}.
Despite these advancements, the domain of mathematical multimodal reasoning such as interpreting charts remains relatively unexplored. Mathematical reasoning in question answering poses unique challenges, as models proficient in natural images struggle with specific types, such as charts. Charts, with their unique formats and the need for logical interpretation, necessitate a different approach to learning compared to conventional VQA models targeting natural images.

Prior chart reasoning methods mainly rely on heuristic rule-based systems, and thus they are not only limited to pre-defined chart formats but also struggle with novel chart types without additional rule formulation \citep{luo2021chartocr}. Moreover, the performance of models utilizing OCR or key-point detection modules are highly dependent on the performance of these modules, and also face significant annotation costs and are typically unable to perform end-to-end reasoning \citep{methani2020plotqa, poco2017reverse}. In response to these limitations, recent approaches have adopted vision-language models trained in an end-to-end manner without heuristic rules \citep{cheng2023chartreader, liu2022matcha}, which, however, require fine-tuning for each specific downstream task, limiting their flexibility.

% \looseness=-1
A novel approach, called Deplot \citep{liu2022deplot}, that combines vision-language models with Large Language Models (LLMs) has emerged as an approach to addressing these issues. 
Specifically, Deplot first transforms charts into tables (i.e., chart-to-table extraction), and then employs the extracted tables along with the LLMs for reasoning.
This approach not only aims to solve the inherent problems of previous methodologies, but also leverages the capability of the LLMs to enhance performance in chart question answering. Converting charts into tables before reasoning offers several advantages, including improved interpretability and the ability to achieve high performance in table reasoning, thus facilitates more accurate and precise reasoning compared to a direct image-based QA.

\begin{figure}[!t]  %%% t: top, b: bottom, h: here
\centering
    \includegraphics[width=0.9\linewidth]{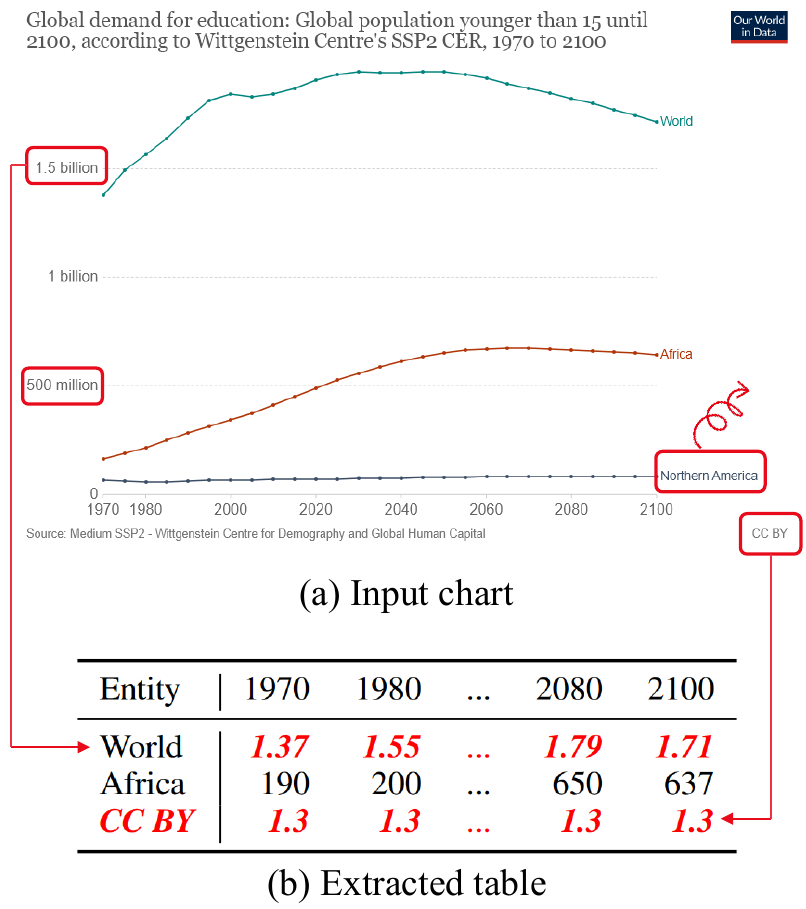} 
    \caption{An example of chart-to-table extraction (from (a) to (b)) by Deplot \citep{liu2022deplot} on ChartQA. 
    % Existing methods fail to incorporate textual information within charts (ex."\textit{million / billion}") and extract unnecessary information from real-world charts (ex."\textit{CC BY}") instead of necessary information (ex. "\textit{Northern America}"), leading to inaccurate reasoning.
    }
    \label{fig:motivation}
    % %{-3.5ex}
\end{figure}

Despite its effectiveness, Deplot suffers from the following two limitations that still need to be addressed.
First, we discovered that Deplot struggles to fully utilize the textual context within a chart for chart-to-table extraction
(See Fig.~\ref{fig:motivation}). 
For example, the counting unit (e.g., ``million'' or ``billion'') is overlooked when converting a chart (Fig.~\ref{fig:motivation}(a)) into a table (Fig.~\ref{fig:motivation}(b)).
Moreover, irrelevant information (e.g., ``CC BY'') is extracted as an entity in the table, while relevant information (e.g., ``Northern America'') is overlooked.
However, as real-world charts often contain information that might not be helpful for chart reasoning (e.g., source credit and ``CC BY''), it is crucial for the model to be able to differentiate between relevant and irrelevant information.
Owing to these challenges, Deplot is prone to eventually extracting inaccurate values within a table, failing in effectively interpreting the information of the chart.

The second limitation of Deplot is that it solely relies on tables to solve chart reasoning tasks, while overlooking the visual information associated with the tables. For this reason, Deplot fails to answer questions regarding the information that cannot be obtained from the extracted table itself (e.g., "What is the value of the third bar from the top?", "What year does the orange line represent?"), indicating a significant shortfall in the model’s capability to interpret visual data.

In this paper, we propose a simple yet effective method, named \proposed, to address the aforementioned limitations of Deplot.
To handle the first limitation, we develop two methods aimed at providing explicit supervision to the model. The first method, named row-col rendering, involves explicitly inserting information about the rows and columns that an extracted table should contain, allowing the model to perceive more accurate textual context. The second method involves conveying only the essential information in the chart to the model, which is built on our observation that converting an original chart containing irrelevant information into a simple chart containing only the essential information significantly improves the performance of table extraction.

To address the second limitation, we present a novel prompt named \prompt~while leveraging a Large Multimodal Model (LMM) to utilize the visual attributes of a chart. Since more advanced reasoning is required when interpreting charts, LMM requires a prompt that is specifically designed for the chart reasoning task, differentiated from natural images. Hence, we provide instructions to the model in a way that is similar to how humans interpret charts for precise reasoning.

Our contributions are summarized as follows:
\begin{enumerate}
    \item We provide guidelines for utilizing textual information within charts for reasoning that was previously overlooked.
    \item \proposed~extracts only essential information from complex charts, preventing irrelevant information from entering the model, resulting in detailed reasoning.
    \item We present a prompt specifically designed for chart reasoning. Through this prompt, LMMs perform more accurate reasoning by mimicking how humans interpret charts.
    \item Extensive experiments validate that~\proposed~successfully addresses the limitations of Deplot. A further appeal of~\proposed~is that it is model-agnostic, i.e., it can be applied to any existing model that involves chart-to-table extraction for chart reasoning beyond Deplot.
\end{enumerate}

\section{Proposed Method:~\proposed}
\label{sec:method}

\begin{figure*}[!t]  %%% t: top, b: bottom, h: here
\centering
    \includegraphics[width=0.95\linewidth]{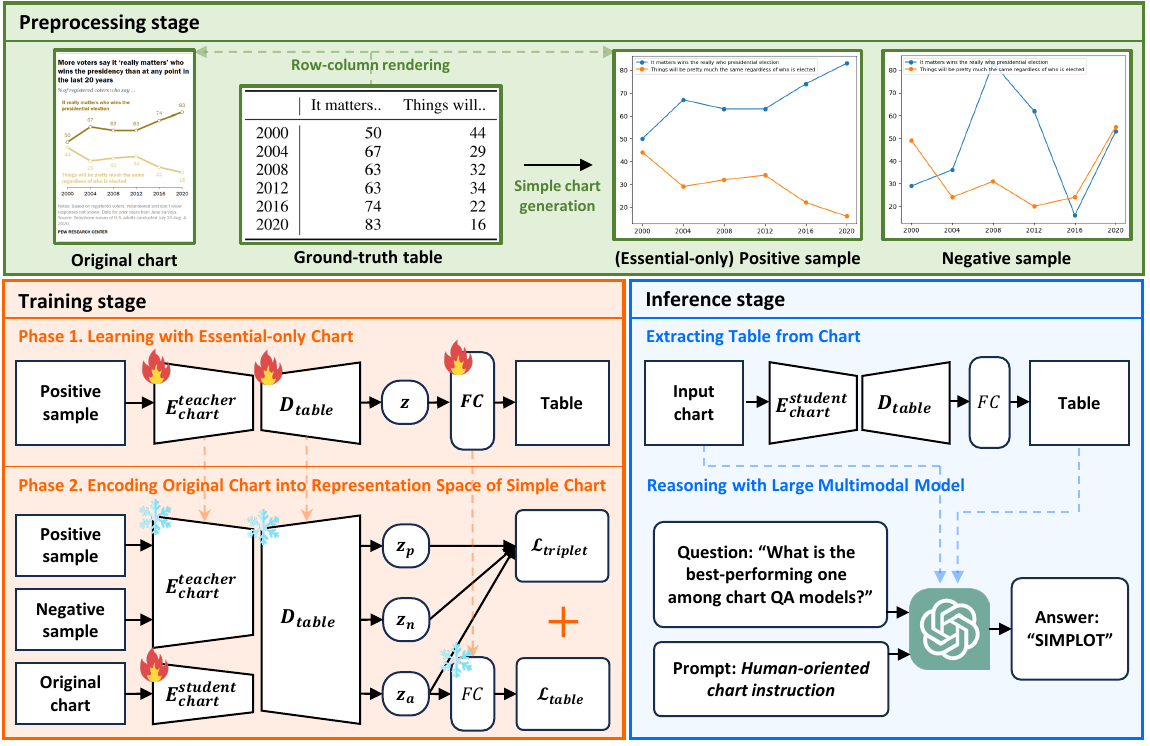} 
    \caption{Overall framework of \proposed. \textcolor{teal}{Upper box} presents the preprocessing stage, which involves generating a simple positive sample containing only essential information from the original chart, as well as a negative sample, along with row and column rendering. \textcolor{orange}{Lower left box} illustrates the training stage including two phases. In Phase 1 of the training stage, a teacher encoder and a table decoder are trained using a simple chart, and in Phase 2, a student encoder is trained with the original chart, while being distilled the knowledge of the teacher encoder on how to generate a table from a simple chart. \textcolor{blue}{Lower right box} illustrates the inference stage, where an LMM receives the original chart and the extracted table along with prompts for reasoning.}
    \label{fig:architecture}
    % %{-3ex}
\end{figure*}

In this section, we describe our proposed method, \proposed. Fig~\ref{fig:architecture} presents the overall framework of \proposed, where \proposed~eventually performs reasoning with an LMM given a table extracted from a chart (Inference stage). In this work, we focus on chart-to-table extraction (Training stage) as a high-quality table enables precise reasoning with a powerful LMM.
It is important to note that \proposed~can be combined with diverse LMM variants, and various techniques such as Chain-of-Thought (CoT) \citep{wei2022chain} can further improve the accuracy of the chart reasoning, which we leave as future work.

\subsection{Preprocessing Stage}\label{sec:preprocess}

Before the training stage, we conduct two preprocessing steps, i.e., 1) Simple Chart Generation, and 2) Row-Column Rendering. Note that as these steps can be readily done offline, they do not increase the training time.

\smallskip
\noindent\textbf{1) Simple Chart Generation. }
We generate a simple chart by excluding irrelevant information from the original chart, keeping only essential elements required for reasoning.
Specifically, since each of the original charts in the dataset is annotated with a table in the CSV format, we use a Python library (i.e., Matplotlib) to plot a chart based on the table. The chart generated in this manner is considered as a simplified version of the original chart, i.e., simple chart.
This process requires no separate training or additional costs and it can be executed offline by running a simple code snippet, so it is scalable and generalizable across various datasets.

\smallskip
\noindent\textbf{2) Row-Column Rendering. }
Inspired by rendering questions over images used for QA \citep{lee2023pix2struct}, we aim to improve the table extraction accuracy by rendering information about the rows and columns that should be included in the table onto the image, enabling the model to utilize this information to extract more accurate tables when converting charts to tables.
Note that in the training stage, since we are given the ground-truth chart-table pairs, we simply render rows and columns of each table onto its paired image containing the chart. 
However, it would not be feasible in the inference stage, since we would be only given the charts without tables. Instead, in the inference stage, 
we utilize a LMM to extract rows and columns from the chart and render them onto the image. Despite the relative low performance of LMM such as GPT-4 \citep{achiam2023gpt} in chart reasoning, we found they can accurately extract row and column information from charts, since such information is given as text and structured in a relatively simple manner. Detailed description of the process is presented in Appendix \ref{app:rendering}.

\subsection{Training Stage: Chart-to-Table Extraction}
\label{sec:charttotable}
Our proposed chart-to-table extraction approach consists of two phases: 
\textbf{Phase 1)} Training a teacher encoder and a table decoder by performing the chart-to-table extraction given simple charts containing only the essential information for reasoning, rather than complex original charts; 
\textbf{Phase 2)} Training a student encoder by extracting the table given the original chart, while being distilled the knowledge by teacher encoder to embed original charts to the embedding space of simple charts.
\subsubsection{Phase 1: Learning with Essential Part from Simple Chart}
\label{sec:phase1}

\looseness=-1
Our model learns the process of extracting tables from previously generated simple charts. 
Specifically, we use Deplot \cite{liu2022deplot} as our backbone model, which is our baseline model that consists of an image encoder and a text decoder, and fine-tune it on the generated simple chart-table pairs. This facilitates the image encoder to obtain representations containing only essential information within the chart, while the text decoder converts this representation into a table format. In this paper, we name the encoder and decoder as \textit{chart encoder} $E_{chart}$ and \textit{table decoder} $D_{table}$, respectively.

\begin{figure}[t]  %%% t: top, b: bottom, h: here
\centering
    %{-1ex}
    \includegraphics[width=0.8\linewidth]{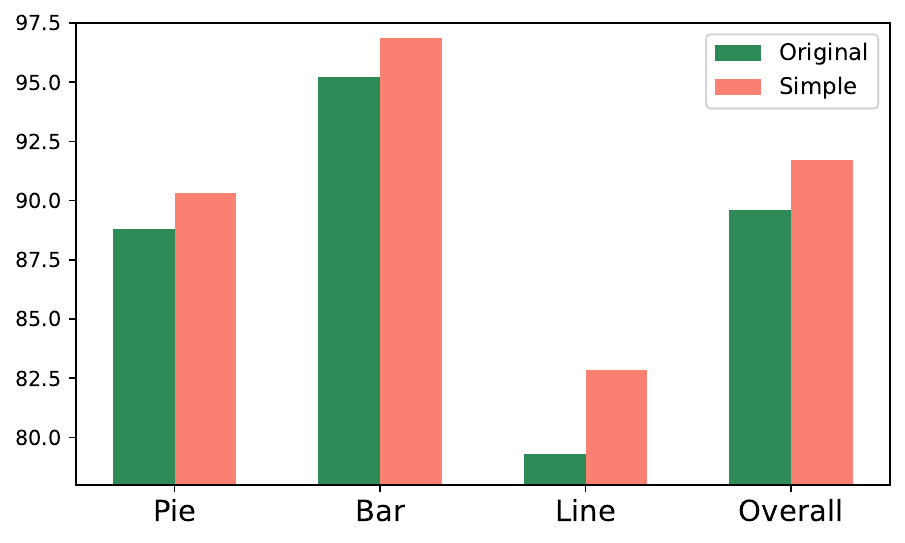} 
    %{-1ex}
    \caption{Accuracy of chart-to-table extraction using the original (in green) and simple charts (in red).}
    \label{fig:simple}
    %{-3ex}
\end{figure}

To corroborate our model design of focusing on the essential parts when learning from charts, in Fig.~\ref{fig:simple}, we compare the chart-to-table extraction performance when using either one of the two chart types (i.e., original chart (in green), and simple chart containing essential information only (in red)) for both training and inference. We observe that using simple charts for chart-to-table extraction greatly outperforms the case when original charts are used.

Note that the chart encoder trained here serves as the teacher encoder $E_{chart}^{teacher}$, providing guidelines for the representation that the student encoder $E_{chart}^{student}$ should learn in the subsequent learning process to be described in Sec. \ref{sec:phase2}. 
Following Deplot, we adopt ViT~\citep{dosovitskiy2020image} as the chart encoders. Note that the trained table decoder $D_{table}$ and a fully connected layer ($FC$) remain frozen in the next stage.

\subsubsection{Phase 2: Encoding Original Chart into Representation Space of Simple Chart}
\label{sec:phase2}
As shown in Fig.~\ref{fig:architecture}, after training the teacher encoder $E_{chart}^{teacher}$ with simple charts as inputs, our goal is to train a student encoder $E_{chart}^{student}$ in a way that the representation of an original chart image obtained from $E_{chart}^{student}$ closely matches the representation of its corresponding simplified chart obtained from $E_{chart}^{teacher}$.
This is to enable $E_{chart}^{student}$ to encode any original chart into the representation space of simplified chart images, thereby improving the accuracy of table generation.
Then, given the chart representations, we generate table representations ${z}$ using the frozen table decoder $D_{table}$ trained in Phase 1.

More precisely, given an original chart $A$ and the simplified chart $P$, we obtain the representations of the original chart $z_a$ and its corresponding sample $z_p$ from sequentially passing into chart encoders and the table decoder. For efficient learning, we define the triplet loss by adding a negative sample $N$, which is created by randomly shuffling the values of the simple chart $P$ and its representation $z_n$ as follows:
\begin{equation}\label{eq:image}
\small
\begin{split}
    \mathcal{L}_{triplet}(A,P,N) = max\{&d(z_a, z_p) 
    - d(z_a,z_n) + m, 0\},\nonumber
\end{split}
\end{equation}
\begin{equation}
\small
\begin{array}{cl}
     & z_a=D_{table}(E_{chart}^{student}(A)), \\ [1ex] 
\text{where}   & z_p=D_{table}(E_{chart}^{teacher}(P)), \\  [1ex]
    & z_n=D_{table}(E_{chart}^{teacher}(N)).
\end{array}
\end{equation}
where $d$ denotes the distance defined as $d(z_i, z_j) = \|z_i - z_j\|_2$, and $m$ denotes the margin which induces the anchor $z_a$ to be closer to the positive sample $z_p$ and farther from the negative sample $z_n$. The purpose of utilizing negative samples is to more precisely discriminate between data that may be mapped closely in the representation space because the charts are similar, and further details and effectiveness are presented in Appendix \ref{app:negative}.

Through this process, the student chart encoder $E_{chart}^{student}$ is trained to extract only the essential information from original charts, even if they contain information that is irrelevant to chart reasoning. The representation $z_a$ obtained after passing the original chart $A$ through the student encoder $E_{chart}^{student}$ and the table decoder $D_{table}$ is then fed into the frozen fully connected layer ($FC$) trained in Phase 1 to generate a table.

Furthermore, since our goal is not just to train $E_{chart}^{student}$ to mimic the representation of simple charts, but to eventually generate tables effectively, we also introduce a cross-entropy loss for table generation as follows:
\begin{equation}\label{eq:table}
\small
%{-1ex}
    T = [\hat{y}_1, \ldots, \hat{y}_n] = FC(z_a),
%{-0.5ex}
\end{equation}
\begin{equation}\label{eq:loss_table}
\small
%{-1ex}
    % \mathcal{L}_{table} = \sum{\log\frac{\exp(x_i)}{\sum\exp(x_i)}}
    \mathcal{L}_{table} = -\frac{1}{N}\sum_{i=1}^{n}\sum_{c=1}^{C} y_{i,c} \log \left( \frac{\exp(\hat{y}_{i,c})}{\sum_{j=1}^{C} \exp(\hat{y}_{i,j})} \right),
%{-1ex}
\end{equation}
where $T$ a linearized textual sequence of the generated table, $n$ and $C$ denote the length of the linearized textual sequence of the table $T$, and the number of classes, respectively, and $y_{i}$ and $\hat{y}_{i}$ denote the $i$-th ground-truth and the predicted token each belonging to one of the $C$ classes, i.e., $y_{i,c}=0$ if $y_i$ belongs to class $c$ and 0 otherwise.
Note that, as in Deplot, lines of $T$ are separated by `\verb|<0x0A>|' (line break), and cells are distinguished by `|'.

The final loss function for chart-to-table extraction is defined as follows: 
\begin{equation}\label{eq:loss_final}
\small
%{-1ex}
% \resizebox{.9\hsize}{!}{
    \mathcal{L}_{final}=\lambda\mathcal{L}_{triplet} + (1-\lambda)\mathcal{L}_{table}.
    % }
%{-1ex}
\end{equation}
where $\lambda$ is a hyperparameter balancing the two losses, i.e., $\mathcal{L}_{triplet}$ and $\mathcal{L}_{table}$.
Note that considering the scale of the two losses, we set $\lambda$ to 0.1 to balance between them. For detailed analysis of hyperparameter $\lambda$, please refer to Appendix \ref{app:lambda}.

In summary, in Phase 2, we align the representation space of $E_{chart}^{student}$, which is trained with original charts, and $E_{chart}^{teacher}$, which is trained with simple charts, thereby allowing $E_{chart}^{student}$ to generate chart representations that mainly contain essential information for chart reasoning. Based on the representations obtained from $E_{chart}^{student}$, we use the frozen decoder $D_{table}$ to generate tables that will be passed on to the LMM in the inference stage.

\subsection{Inference Stage: Reasoning with Extracted Table}
\label{sec:reasoning}

In the inference stage, we use a Large Multimodal Model (LMM) to perform various reasoning tasks given the tables obtained from the training stage as input.
However, relying solely on tables without the associated visual information to solve chart reasoning tasks makes it impossible for the model to answer questions regarding visual information in charts (e.g., "What is the value of the third bar from the top?", "What year does the orange line represent?").
To address this issue, we additionally provide the original chart as another input to the LMM to enable the model to provide answers to a wider range of questions. 
By doing so, we expect the model to not only answer to questions regarding precise chart values by referring to the tables, but also to answer about visual attributes not contained in the table by referring to the original chart. Please refer to Appendix \ref{app:limitation} for experimental results on the effects of using the table and original chart together.

\begin{figure}[!b]  %%% t: top, b: bottom, h: here
\centering
    %{-1ex}
    \includegraphics[width=0.75\linewidth]{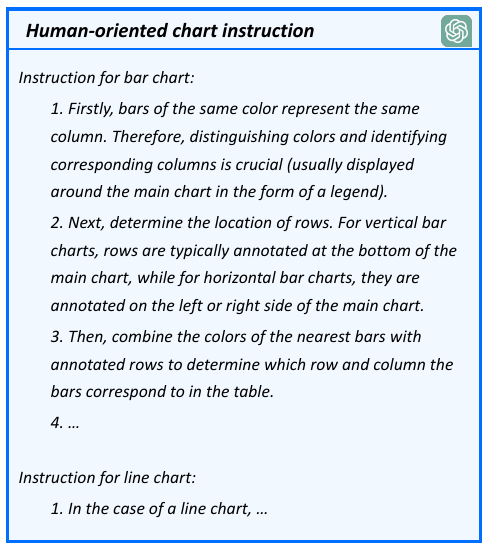} 
    %{-1ex}
    \caption{A snippet of the proposed prompt.}
    \label{fig:miniprompt}
%{-3ex}
\end{figure}

\smallskip
\noindent\textbf{\prompt. }
Furthermore, we present a novel prompt that is specifically designed for chart reasoning, named \prompt. 
Although prompt engineering is a widely studied topic for processing natural images \citep{kim2023llm4sgg, li2022blip}, it is crucial to develop specialized prompts specifically tailored  for charts, due to the inherent nature of chart reasoning tasks, i.e., they require more advanced advanced thinking steps for interpretation.
Our proposed prompt provides instructions that mimic how humans interpret charts, enabling more accurate reasoning within the LMM and its effectiveness will be demonstrated in Section \ref{sec:overallresult}. We provide a snippet of our proposed prompt in Fig.~\ref{fig:miniprompt}. For detailed description of \prompt, please refer to Appendix \ref{app:prompt}.

\section{Experiments}
\label{sec:experiments}

\textbf{Dataset.}
In this paper, we evaluate \proposed~using widely used dataset for chart reasoning, ChartQA \citep{masry2022chartqa} comprising three types of charts (i.e., pie, bar, and line), and PlotQA \citep{methani2020plotqa} including three types of charts (i.e., dot line, line, bar). ChartQA also includes human-authored and LLM-augmented QA pairs, while PlotQA consists of QA pairs generated from templates created by manually analyzed sample questions collected through crowd-sourcing. Detailed explanation for dataset is presented in Appendix \ref{app:dataset}.

\smallskip
\noindent\textbf{Baselines.}
In this paper, we use various models as baselines that can perform chart reasoning. The baselines we employ are divided into three categories including Vision-Language pre-trained models (VLP), fully supervised models for chart reasoning, and models which extract table and perform reasoning using LLMs. For a fair comparison, we employ GPT-4V for all methods utilizing extracted table. Detailed descriptions of the models are provided in the Appendix \ref{app:baseline}.

\smallskip
\noindent\textbf{Evaluation protocol.}
For question answering based on charts, we report Relaxed Accuracy (RA) of 2,500 questions in the test set following previous works \citep{lee2023pix2struct, kantharaj2022chart, masry2022chartqa, kantharaj2022opencqa, liu2022matcha}.

Furthermore, we propose a proper metric to evaluate the chart-to-table extraction performance, employing a slightly modified metric derived from Relative Mapping Similarity ($RMS$), presented by Deplot \citep{liu2022deplot}. We discover that the vanilla $RMS$ does not accurately measure the chart-to-table performance in certain cases, and thus we introduce a metric named Relative Distance ($RD$). For the differences between $RMS$ and $RD$, please refer to the Appendix \ref{app:metric}.

\subsection{Experimental Results}
\label{sec:result}

\subsubsection{Results on Chart-to-Table Extraction}\label{sec:chart2table}

In Table \ref{tab:charttotable}, we observe that \proposed~exhibits superior chart-to-table extraction performance compared to existing methods including Deplot \citep{liu2022deplot} and UniChart \citep{masry2023unichart}. We argue that the superiority of \proposed~in terms of chart-to-table extraction eventually leads to more precise reasoning as will be shown in Section \ref{sec:overallresult}. 
While the difference in chart-to-table performance may seem small, we argue that even minor errors such as slightly incorrect values or column and row names during the chart-to-table extraction may entail completely incorrect answers, as shown in Table \ref{tab:main}.

We also have mentioned that GPT-4V can achieve more accurate text extraction rather than numerical extraction, which is why we utilize it for row and column extraction. Nevertheless, to further validate the necessity of a model specifically targeting charts, we also conduct table extraction (both text and numerical information) using GPT-4V, and the results are presented in the Table \ref{tab:charttotable}.
As previously argued, GPT-4V is well-suited for extracting text information from charts, so it demonstrated comparable performance on pie charts where values and rows/columns are primarily given as text. However, for line charts, where most values must be inferred from the image rather than being provided as text, GPT-4V showed significantly lower performance compared to other models.

This explains the low QA performance in Table \ref{tab:main} as well, and conducting QA using tables extracted by GPT-4V is inefficient both in terms of performance and cost. Furthermore, these results support our argument that existing vision-language models still lack effectiveness when dealing with chart images while performing well on natural images, so methods specifically trained to target charts are necessary.

\begin{table}[t]
\centering
\resizebox{0.75\linewidth}{!}{
\begin{tabular}{l|c|c|c|c}
\toprule
\multirow{2}{*}{Models} & \multicolumn{3}{c|}{Chart type} & \multirow{2}{*}{Overall} \\
\cline{2-4}
& Pie & Bar & Line &  \\
\midrule
GPT-4V & \underline{90.13} & 91.53 & 71.51 & 84.24 \\
UniChart & 84.86 & 92.58 & \textbf{85.16} & 88.03 \\
Deplot  & 88.82  & \underline{96.37}  & 82.25 & \underline{90.95} \\
\rowcolor{grey} \textbf{\proposed} &  \textbf{91.41} & \textbf{96.87}  & \underline{84.74} & \textbf{92.32} \\
\bottomrule
\end{tabular}
 }
\caption{Chart-to-table extraction performance ($RD_{F1}$) on the ChartQA dataset over various chart types.}
\label{tab:charttotable}
\vspace{-1.5ex}
\end{table}

\subsubsection{Results on Chart Reasoning} \label{sec:overallresult}

\begin{table}[t]
\centering
\resizebox{0.9\linewidth}{!}{
\begin{tabular}{cl|c|c|c}
\cline{2-5}
& \multirow{2}{*}{Models} & \multicolumn{3}{c}{Data type}  \\
\cline{3-5}
& & Human & Augmented & Overall \\
\cline{2-5}
\multirow{8}{*}{\rotatebox[origin=c]{90}{\bf VLP models}}
& TaPas & 28.72 & 53.84 & 41.28 \\
& V-TaPas & 29.60  & 61.44 & 45.52 \\
& T5 & 25.12 & 56.96 & 41.04 \\
& VL-T5 & 26.24 & 56.88 & 41.56 \\
& PaLI & 30.40 & 64.90 & 47.65 \\
& Mini-GPT & 8.40 & 15.60 & 12.00 \\
& LLaVa & 37.68 & 72.96 & 55.32 \\
& GPT-4V & 56.48 & 63.04 & 59.76 \\
\cline{2-5}
\multirow{8}{*}{\rotatebox[origin=c]{90}{\bf Supervised}}
& ChartQA & 40.08 & 63.60 & 51.84 \\
& ChartT5 & 31.80 & 74.40 & 53.10 \\
& Pix2Struct & 30.50 & 81.60 & 56.05 \\
& MatCha & 38.20 & 90.20  & 64.20 \\
& Unichart & 43.92 & 88.56 & 66.24 \\ 
& ChartLlama & 48.96 & \underline{90.36} & 69.66 \\
& ChartAssisstant & 65.90 & \textbf{93.90} & \underline{79.90} \\
& ChartInstruct & 45.52 & 87.76 & 66.64\\
% & ChartGemma & \underline{69.52} & \underline{90.80} & \underline{80.16} \\
\cline{2-5}
\multirow{3}{*}{\rotatebox[origin=c]{90}{\bf Table}}
& Deplot & 62.71 & 78.63 & 70.67 \\
% & Deplot + img. & 72.39 & 85.01 & 78.70 \\
& Unichart\footnotemark  & \underline{67.04} & 69.92  & 68.48 \\
\cline{2-5}
% & \cellcolor{grey} \textbf{\proposed} & \cellcolor{grey}\textbf{73.91} & \cellcolor{grey}\textbf{85.67} & \cellcolor{grey}\textbf{79.79} \\
& \cellcolor{grey} \textbf{\proposed} & \cellcolor{grey}\textbf{78.07} & \cellcolor{grey}88.42 & \cellcolor{grey}\textbf{83.24} \\
\cline{2-5}
\end{tabular}
}
\caption{Chart question answering performance (RA) on the ChartQA dataset.}
\vspace{-2.5ex}
\label{tab:main}
\end{table}
\footnotetext{The authors also provide pre-trained model for chart to table extraction. We use this model to extract the table and conduct reasoning in the same setting of Deplot and \proposed.}

The performance of question answering on the ChartQA dataset is summarized in Table \ref{tab:main}. 
1) In general, we observe that methods designed with a focus on chart reasoning (i.e., `Supervised' and `Table') outperform VLP in QA tasks. 
This demonstrates that despite the superior performance of vision-language models in performing a wide range of tasks in the traditional natural image domain, they face difficulties in interpreting charts due to the significantly different characteristics of natural images and charts.
2) Furthermore, noticeable performance difference is observed among models aimed at chart reasoning. Methods using tables such as Deplot, Unichart, and \proposed, generally outperform models that conduct reasoning based solely on images (i.e., `Supervised'). This suggests that going through tables allows for more detailed reasoning, aligning well with our motivation that emphasizes the importance of effectively extracting tables. Additionally, models utilizing extracted tables and LMM can efficiently handle various tasks such as OpenCQA \citep{kantharaj2022opencqa} without any additional data and training. 3) Among the table-based reasoning models, \proposed~performs the best by utilizing textual information, which is essential for obtaining the most accurate table, and by employing prompts mimicking the human reasoning process. Additional experimental results on PlotQA dataset and effectiveness of our method are presented in Appendix \ref{app:plotqaexperiment} and \ref{sec:furtheranalysis}, respectively.

\smallskip
\noindent\textbf{Discussion.} 
The overall performance of \proposed~is consistently higher than other baselines; however, on the augmented set, it sometimes exhibits slightly lower performance compared to some models. This can be attributed to the characteristic of the augmented set, which is obtained by fine-tuning the T5 model on the SQuAD QA dataset \citep{rajpurkar2016squad} to generate questions and answers. Consequently, the questions in the augmented set are mostly simpler and relatively easier to answer compared to those in the human set. Models that are fully fine-tuned on such questions tend to show notably higher performance on augmented data. For instance, MatCha \citep{liu2022matcha} shows a performance of 90.2 on the augmented set, but its performance drops significantly to 38.2 on the human set which is composed of more complex QA pairs, resulting in an overall performance of 64.2. In contrast, \proposed's high performance even on the human set without additional fine-tuning suggests that proposed method is effective for more complex reasoning tasks.

\subsection{Further Analysis}\label{sec:furtheranalysis}
\noindent\textbf{Ablation Study. }
We conduct ablation studies to determine the impact of each module of~\proposed~on the performance. Upper part of the Table \ref{tab:ablation} shows the effects of the two modules we introduced for chart-to-table extraction: 1) row-column rendering and 2) distillation from a simple chart. These results demonstrate that each module enhances the performance of table extraction, thereby proving the effectiveness of our design. 

Additionally, to verify the effect of the proposed prompt, we also present the QA performance with and without the \prompt~(lower part of Table \ref{tab:ablation}). We observe that our prompt that mimics how humans think enhances the performance of the QA task, which confirms the importance of using appropriate prompts tailored to the task when employing LMMs.
It is notable that providing more diverse in-context examples could lead to even greater performance gains.

\begin{table}[!h]
  \centering
  \resizebox{0.85\columnwidth}{!}{
  \begin{tabular}{ccc|cc}
    \toprule
     Row-col rendering & Simple chart & Prompt &  $RD_{F1}$ & $RA$ \\
     \midrule
     \ding{55} & \ding{55}  & - & 90.95 & - \\
     \ding{51} & \ding{55}  &  - &91.40 & -  \\
     \ding{55} & \ding{51}  &  - &91.86  & -   \\
     \ding{51} & \ding{51}  &  - &\textbf{92.32} & -  \\ 
     \midrule
     - &  - &  \ding{55} & - & 79.79  \\
     - &  - &  \ding{51} & - & \textbf{83.24}   \\
  \bottomrule
\end{tabular}
}
\caption{Ablation studies on each component of~\proposed~for chart-to-table extraction and QA tasks.}
\label{tab:ablation}
\vspace{-1ex}
\end{table}

\begin{figure}[!t]  %%% t: top, b: bottom, h: here
\centering
    %{-1ex}
    \includegraphics[width=0.98\linewidth]{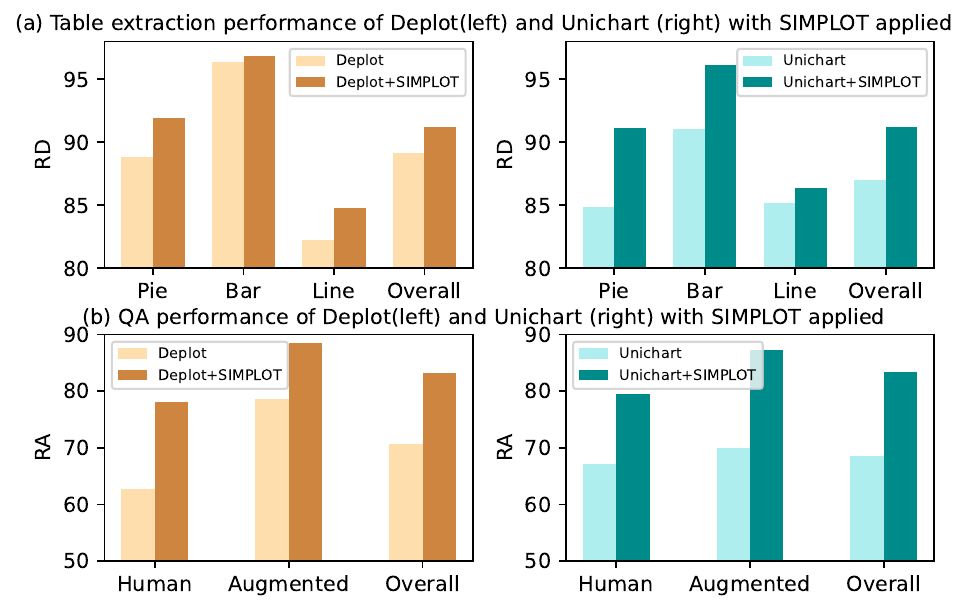} 
    %{-2ex}
    \caption{Improvements in performance observed when \proposed~is applied across diverse models.}
    \label{fig:adaptation}
    \vspace{-2ex}
\end{figure}

\smallskip
\noindent\textbf{\proposed~is Model-agnostic. }
While \proposed~is originally designed to address the limitations of Deplot, it is model-agnostic, i.e., \proposed~can be applied to any existing table-based reasoning model to enhance its performance. 
Here, we applied~\proposed~to one of our baselines, Unichart \citep{masry2023unichart}, which is pre-trained to perform chart-to-table extraction task. 
Specifically, Fig. \ref{fig:adaptation} (a) demonstrates significant performance enhancements with \proposed~applied to Deplot and Unichart, respectively. Consequently, Fig. \ref{fig:adaptation} (b) show a substantial increase in question answering performance. These results verifies the generality and practicality of~\proposed.

\smallskip
\noindent\textbf{Using Images as Input to Unichart/Deplot for Inference. }
\looseness=-1
Here, we further demonstrate that using both tables and images is essential for chart reasoning, as emphasized earlier. Table \ref{tab:furtheranalysis} shows that the performance of the vanilla Deplot and the vanilla Unichart, both of which only use tables for reasoning, is improved when images are also utilized during inference. This indicates a clear drawback of relying solely on tables and proves the necessity of using images with specifically designed prompts. Moreover, even without the prompt, \proposed~outperforms the baselines with images used during inference, thanks to the superior table extraction capability of~\proposed.

\begin{table}[h]
\small
\centering
\resizebox{1\linewidth}{!}{

\begin{tabular}{l|ccc}
\toprule

Models & Human & Augmented & Overall  \\
\midrule
Unichart & 67.04 & 69.92  & 68.48 \\
Unichart + img. & 75.04 & \textbf{88.82} & 81.93 \\
Unichart + \proposed~w/o prompt & \underline{76.56} & \underline{88.64} & \underline{82.60} \\
Unichart + \proposed & \textbf{79.56} & 87.18 & \textbf{83.37} \\
\midrule
Deplot & 62.71 & 78.63 & 70.67 \\
Deplot + img. & 72.39 & 85.01 & 78.70 \\
Deplot + \proposed~w/o prompt & \underline{73.91} & \underline{85.67} & \underline{79.79} \\
Deplot + \proposed & \textbf{78.07} & \textbf{88.42} & \textbf{83.24} \\
\bottomrule
\end{tabular}
}
\caption{Chart question answering performance (RA) with/without image on ChartQA dataset.}
\label{tab:furtheranalysis}
%{-2ex}
\end{table}

\smallskip
\noindent\textbf{Broad Applicability of the Proposed Prompt. }
To verify the effectiveness of our proposed prompt, we apply it to Deplot with images (upper part of Table \ref{tab:prompteffect_main}). We observe that applying the proposed prompt to the `Deplot+img' indeed greatly enhances its performance.

\smallskip
\noindent\textbf{Performance on more Challenging Settings. }
It is worth noting that `Deplot+img+prompt' now performs competitively to~\proposed. However, we argue that this is mainly due to the simplicity of the ChartQA dataset, which makes the dataset insufficient to evaluate the table extraction capability of \proposed. For example, ChartQA contains questions that can be answered by referencing just one row of the extracted table. 

To evaluate the models with more challenging questions, we randomly sample 100 test images and then provide GPT-4 with the following instruction: \textit{"Create a challenging question-answer pair that requires referencing at least two rows and two columns to solve."}. Through this process, we generated 100 question-answer pairs and eventually obtained 85 pairs after manually filtering out inaccurate pairs, and then evaluate `Deplot+img+prompt' and~\proposed~on these question-answer pairs (lower part of Table \ref{tab:prompteffect_main}). We observe that the performance gap greatly increases as the questions get more challenging, which demonstrates the importance of extracting accurate tables when encountering challenging questions. For the specific example of the challenging question and how the extracted tables from \proposed~and Deplot lead to differences in complex QA performance, please refer to Appendix \ref{app:challenging}.

\begin{table}[!h]
\centering
\resizebox{0.95\linewidth}{!}{
\begin{tabular}{cl|ccc}
\cline{2-5}
& Models & Human & Augmented & Overall \\
\cline{2-5}
\multirow{3}{*}{\rotatebox[origin=c]{90}{\bf Easy}}
& Deplot + img. & 72.39 & 85.01 & 78.70 \\
& Deplot + img. + prompt & 77.75 & 88.30 & 83.03 \\
& \proposed & \textbf{78.07}  & \textbf{88.42} & \textbf{83.24} \\
\cline{2-5}
\multirow{2}{*}{\rotatebox[origin=c]{90}{\bf Hard}}
& Deplot + img. + prompt & - & - & 49.41 \\
& \proposed & - & - & \textbf{65.88} \\
\cline{2-5}
\end{tabular}
}
\caption{Performance on more challenging questions.}
\label{tab:prompteffect_main}
% \vspace{-1ex}
\end{table}

\begin{table}[!h]
\centering
% \vspace{-1ex}
\resizebox{1\linewidth}{!}{
\begin{tabular}{l|ccc}
\cline{1-4}
Models & Com. (Easy) & Com. (Hard) & Sequential \\
\cline{1-4}
Deplot+img.+prompt & 57.50 & 50.00 & 26.80 \\
\proposed & \textbf{59.20}  & \textbf{58.10} & \textbf{36.50} \\
\cline{1-4}
\end{tabular}
}
\caption{Performance on MultiChartQA dataset.}
\label{tab:multichart}
\vspace{-2ex}
\end{table}

Furthermore, questions that necessitate referring to multiple charts \citep{zhu2024multichartqa} are expected to be significantly impacted by the accuracy of table extraction when it comes to improving QA performance. The additional experimental results are presented in the Table \ref{tab:multichart}. On the comparative dataset, which requires simultaneously comparing values from two or more charts, and the sequential dataset, which requires finding new information in one chart based on information from another chart, our proposed method consistently outperforms Deplot. Since these tasks require examining multiple charts simultaneously, the results strongly support our claim that accurate table extraction leads to better performance on complex tasks. Moreover, the comparative dataset is divided into easy and hard categories, and the greater performance differences observed on harder questions indicate that even small improvements in table extraction accuracy can significantly impact accurate QA.

\smallskip
\noindent\textbf{Adaptability of Chart-to-table Extraction to another task. }
As we have argued before, extracting tables and combining them with LMM not only improves the accuracy of QA but also offers the advantage of immediately being applicable to other tasks such as OpenCQA \citep{kantharaj2022opencqa} without additional training.

Table \ref{tab:opencqa} presents the BLEU score \citep{papineni2002bleu} of a model fine-tuned for OpenCQA (ChartInstruct), and those of models that utilize extracted tables (Deplot \& \proposed). Here, we demonstrate that leveraging tables allows for effective performance on other tasks without additional training, while also proving that accurately extracting tables enhances accuracy when performing these tasks. Note that this dataset includes charts in a different format from those used to train \proposed, so training with the charts from this dataset could potentially improve performance.

\begin{table}[h]
\small
\centering
\resizebox{0.6\linewidth}{!}{
\begin{tabular}{l|c}
\toprule
Models & BLEU score  \\
\midrule
Gemini & 6.84  \\
GPT-4V & 3.31 \\
ChartInstruct  & 5.27 \\
Deplot & 6.88 \\
\proposed & \textbf{7.11}  \\
\bottomrule
\end{tabular}
}
\caption{Open-ended question answering performance on OpenCQA dataset.}
\label{tab:opencqa}
\vspace{-1ex}
\end{table}

\begin{figure}[!t]  %%% t: top, b: bottom, h: here
\centering
    %{-1ex}
    \includegraphics[width=1\linewidth]{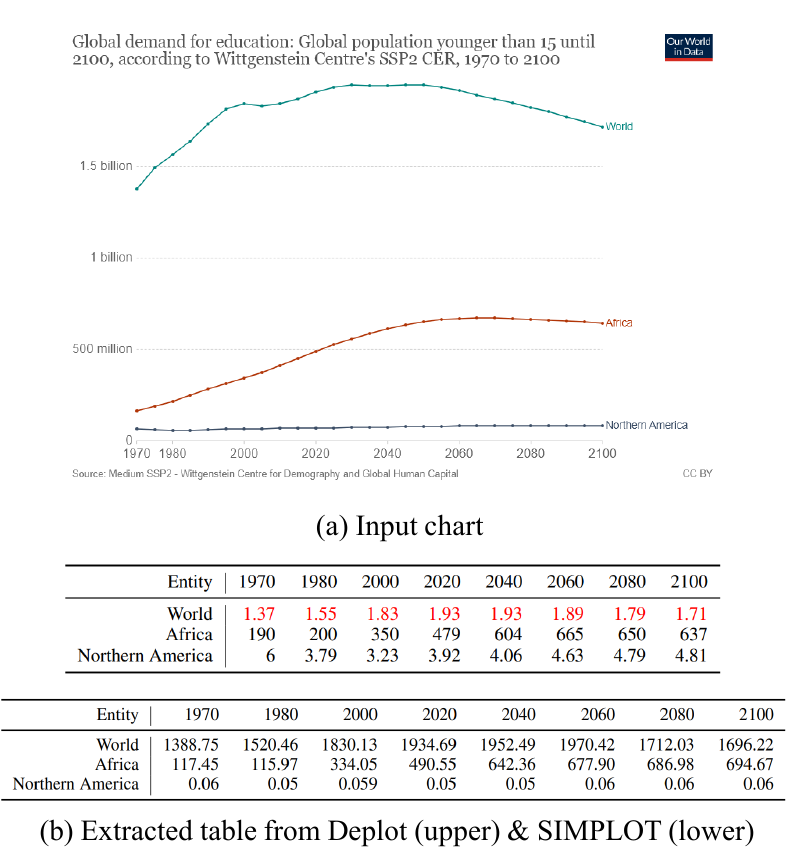} 
    %{-4ex}
    \caption{\proposed~captures contextual information (\textit{billion/million}) and extract precise values in the chart.}
    \label{fig:case1}
    %{-3ex}
\end{figure}

\smallskip
\noindent\textbf{Case Studies. }
\looseness=-1
We verify whether~\proposed~correctly captures textual information within charts. Fig. \ref{fig:case1}~illustrates that the table extracted by Deplot fails to consider the counting unit, i.e., "million/billion," present in the chart, while \proposed~correctly captures the textual information.

Please refer to Appendix \ref{app:casestudy} for further case studies including the results that demonstrate the effectiveness of proposed prompt and Appendix \ref{app:error} for error analysis.

\section{Related Work}
\label{sec:relatedwork}

Early chart processing methods relied on rule-based approaches, which lacked flexibility for various chart types \citep{balaji2018chart, choi2019visualizing}. OCR and key-point detection methods are utilized to address these limitations, but they depend on data annotation and module performance, making them time-consuming \citep{poco2017reverse, zhou2023enhanced, xue2023chartdetr}. 

Recent papers train end-to-end vision-language models to improve chart reasoning without heuristic rules, but they need fine-tuning for specific tasks \citep{meng2024chartassisstant, huang2023lvlms, masry2023unichart, han2023chartllama}, so recent approaches transform charts into tables and utilized LLM for question answering while enhancing interpretability and allowing the model apply to various downstream task without additional training \citep{liu2022deplot, xia2023structchart}. For complete related works, please refer to the Appendix \ref{app:relatedwork}.

\section{Conclusion}
\label{sec:conclusion}
We propose \proposed, which extracts only the essential information required for chart reasoning. Furthermore, by leveraging textual information inside the charts, \proposed~enables accurate chart reasoning. We also propose a novel prompt specifically designed for chart reasoning, which guides LMM to imitate how humans interpret chart. Through extensive experiments, we demonstrate that~\proposed~effectively handle concerns raised by existing works while improving performance, and also can be applied to other various tasks.

\clearpage

\section*{Limitations}
As with existing chart reasoning methods, \proposed~is also expected to have limited performance on unseen out-of-distribution (OOD) data. This is a future research area that all chart reasoning methods, including ours, should consider. However, please note that our proposed method can address various downstream tasks with just one round of training to extract tables from OOD charts, unlike other methods that require separate training for each task. Meanwhile, we conduct inference without training for the OpenCQA dataset where no tables were provided, and we observed that \proposed~outperformed the existing baselines, indicating that our proposed method ensures a certain level of generalization performance for OOD charts.

\section*{Ethics Statement}
\proposed~has not been trained with private or sensitive data. This significantly reduces the risk of generating harmful or misleading content. ChartQA and PlotQA dataset, which we utilize to collect chart-table pairs, is publicly accessible for research purposes. To make sure the transparency and reproducibility of our experiments, we provide detailed information on hyperparameter configurations in our paper and publicly share our source code. This careful approach mitigates the potential for unethical outcomes associated with data usage. While our models demonstrate state-of-the-art performance on the both dataset, we acknowledge the possibility of their misuse. There exists a risk that our models could be exploited to mislead the public about the content and implications of charts. Despite our models' high performance, we cannot guarantee their outputs will always be accurate, emphasizing the need for critical interpretation and verification of results.

\section*{Acknowledgements}
This work was supported by Institute of Information \& Communications Technology Planning \& Evaluation(IITP) grant funded by the Korea government(MSIT) (RS-2023-00216011, Development of artificial complex intelligence for conceptually understanding and inferring like human) and National Research Foundation of Korea(NRF) funded by Ministry of Science and ICT (NRF-2022M3J6A1063021).

\bibliography{custom}

\begin{thebibliography}{54}
\expandafter\ifx\csname natexlab\endcsname\relax\def\natexlab#1{#1}\fi

\bibitem[{Achiam et~al.(2023)Achiam, Adler, Agarwal, Ahmad, Akkaya, Aleman, Almeida, Altenschmidt, Altman, Anadkat et~al.}]{achiam2023gpt}
Josh Achiam, Steven Adler, Sandhini Agarwal, Lama Ahmad, Ilge Akkaya, Florencia~Leoni Aleman, Diogo Almeida, Janko Altenschmidt, Sam Altman, Shyamal Anadkat, et~al. 2023.
\newblock Gpt-4 technical report.
\newblock \emph{arXiv preprint arXiv:2303.08774}.

\bibitem[{Alayrac et~al.(2022)Alayrac, Donahue, Luc, Miech, Barr, Hasson, Lenc, Mensch, Millican, Reynolds et~al.}]{alayrac2022flamingo}
Jean-Baptiste Alayrac, Jeff Donahue, Pauline Luc, Antoine Miech, Iain Barr, Yana Hasson, Karel Lenc, Arthur Mensch, Katherine Millican, Malcolm Reynolds, et~al. 2022.
\newblock Flamingo: a visual language model for few-shot learning.
\newblock \emph{Advances in Neural Information Processing Systems}, 35:23716--23736.

\bibitem[{Antol et~al.(2015)Antol, Agrawal, Lu, Mitchell, Batra, Zitnick, and Parikh}]{antol2015vqa}
Stanislaw Antol, Aishwarya Agrawal, Jiasen Lu, Margaret Mitchell, Dhruv Batra, C~Lawrence Zitnick, and Devi Parikh. 2015.
\newblock Vqa: Visual question answering.
\newblock In \emph{Proceedings of the IEEE international conference on computer vision}, pages 2425--2433.

\bibitem[{Balaji et~al.(2018)Balaji, Ramanathan, and Sonathi}]{balaji2018chart}
Abhijit Balaji, Thuvaarakkesh Ramanathan, and Venkateshwarlu Sonathi. 2018.
\newblock Chart-text: A fully automated chart image descriptor.
\newblock \emph{arXiv preprint arXiv:1812.10636}.

\bibitem[{Biten et~al.(2019)Biten, Tito, Mafla, Gomez, Rusinol, Mathew, Jawahar, Valveny, and Karatzas}]{biten2019icdar}
Ali~Furkan Biten, Ruben Tito, Andres Mafla, Lluis Gomez, Mar{\c{c}}al Rusinol, Minesh Mathew, CV~Jawahar, Ernest Valveny, and Dimosthenis Karatzas. 2019.
\newblock Icdar 2019 competition on scene text visual question answering.
\newblock In \emph{2019 International Conference on Document Analysis and Recognition (ICDAR)}, pages 1563--1570. IEEE.

\bibitem[{Bujard et~al.(1987)Bujard, Gentz, Lanzer, Stueber, Mueller, Ibrahimi, Haeuptle, and Dobberstein}]{bujard198726}
Hermann Bujard, Reiner Gentz, Michael Lanzer, Dietrich Stueber, Michael Mueller, Ibrahim Ibrahimi, Marie-Therese Haeuptle, and Bernhard Dobberstein. 1987.
\newblock [26] a t5 promoter-based transcription-translation system for the analysis of proteins in vitro and in vivo.
\newblock In \emph{Methods in enzymology}, volume 155, pages 416--433. Elsevier.

\bibitem[{Chen et~al.(2022{\natexlab{a}})Chen, Ma, Wang, and Cohen}]{chen2022program}
Wenhu Chen, Xueguang Ma, Xinyi Wang, and William~W Cohen. 2022{\natexlab{a}}.
\newblock Program of thoughts prompting: Disentangling computation from reasoning for numerical reasoning tasks.
\newblock \emph{arXiv preprint arXiv:2211.12588}.

\bibitem[{Chen et~al.(2022{\natexlab{b}})Chen, Wang, Changpinyo, Piergiovanni, Padlewski, Salz, Goodman, Grycner, Mustafa, Beyer et~al.}]{chen2022pali}
Xi~Chen, Xiao Wang, Soravit Changpinyo, AJ~Piergiovanni, Piotr Padlewski, Daniel Salz, Sebastian Goodman, Adam Grycner, Basil Mustafa, Lucas Beyer, et~al. 2022{\natexlab{b}}.
\newblock Pali: A jointly-scaled multilingual language-image model.
\newblock \emph{arXiv preprint arXiv:2209.06794}.

\bibitem[{Chen et~al.(2020)Chen, Li, Yu, El~Kholy, Ahmed, Gan, Cheng, and Liu}]{chen2020uniter}
Yen-Chun Chen, Linjie Li, Licheng Yu, Ahmed El~Kholy, Faisal Ahmed, Zhe Gan, Yu~Cheng, and Jingjing Liu. 2020.
\newblock Uniter: Universal image-text representation learning.
\newblock In \emph{European conference on computer vision}, pages 104--120. Springer.

\bibitem[{Cheng et~al.(2023)Cheng, Dai, Li, Sun, Mitamura, and Hauptmann}]{cheng2023chartreader}
Zhi-Qi Cheng, Qi~Dai, Siyao Li, Jingdong Sun, Teruko Mitamura, and Alexander~G Hauptmann. 2023.
\newblock Chartreader: A unified framework for chart derendering and comprehension without heuristic rules.
\newblock \emph{arXiv preprint arXiv:2304.02173}.

\bibitem[{Chiang et~al.(2023)Chiang, Li, Lin, Sheng, Wu, Zhang, Zheng, Zhuang, Zhuang, Gonzalez et~al.}]{chiang2023vicuna}
Wei-Lin Chiang, Zhuohan Li, Zi~Lin, Ying Sheng, Zhanghao Wu, Hao Zhang, Lianmin Zheng, Siyuan Zhuang, Yonghao Zhuang, Joseph~E Gonzalez, et~al. 2023.
\newblock Vicuna: An open-source chatbot impressing gpt-4 with 90\%* chatgpt quality.
\newblock \emph{See https://vicuna. lmsys. org (accessed 14 April 2023)}.

\bibitem[{Cho et~al.(2021)Cho, Lei, Tan, and Bansal}]{cho2021unifying}
Jaemin Cho, Jie Lei, Hao Tan, and Mohit Bansal. 2021.
\newblock Unifying vision-and-language tasks via text generation.
\newblock In \emph{International Conference on Machine Learning}, pages 1931--1942. PMLR.

\bibitem[{Choi et~al.(2019)Choi, Jung, Park, Choo, and Elmqvist}]{choi2019visualizing}
Jinho Choi, Sanghun Jung, Deok~Gun Park, Jaegul Choo, and Niklas Elmqvist. 2019.
\newblock Visualizing for the non-visual: Enabling the visually impaired to use visualization.
\newblock In \emph{Computer Graphics Forum}, volume~38, pages 249--260. Wiley Online Library.

\bibitem[{Dosovitskiy et~al.(2020)Dosovitskiy, Beyer, Kolesnikov, Weissenborn, Zhai, Unterthiner, Dehghani, Minderer, Heigold, Gelly et~al.}]{dosovitskiy2020image}
Alexey Dosovitskiy, Lucas Beyer, Alexander Kolesnikov, Dirk Weissenborn, Xiaohua Zhai, Thomas Unterthiner, Mostafa Dehghani, Matthias Minderer, Georg Heigold, Sylvain Gelly, et~al. 2020.
\newblock An image is worth 16x16 words: Transformers for image recognition at scale.
\newblock \emph{arXiv preprint arXiv:2010.11929}.

\bibitem[{Gard{\`e}res et~al.(2020)Gard{\`e}res, Ziaeefard, Abeloos, and Lecue}]{garderes2020conceptbert}
Fran{\c{c}}ois Gard{\`e}res, Maryam Ziaeefard, Baptiste Abeloos, and Freddy Lecue. 2020.
\newblock Conceptbert: Concept-aware representation for visual question answering.
\newblock In \emph{Findings of the Association for Computational Linguistics: EMNLP 2020}, pages 489--498.

\bibitem[{Han et~al.(2023)Han, Zhang, Chen, Yang, Wang, Yu, Fu, and Zhang}]{han2023chartllama}
Yucheng Han, Chi Zhang, Xin Chen, Xu~Yang, Zhibin Wang, Gang Yu, Bin Fu, and Hanwang Zhang. 2023.
\newblock Chartllama: A multimodal llm for chart understanding and generation.
\newblock \emph{arXiv preprint arXiv:2311.16483}.

\bibitem[{Holmgren et~al.(2012)Holmgren, Davidsson, Persson, and Ramstedt}]{holmgren2012tapas}
Johan Holmgren, Paul Davidsson, Jan~A Persson, and Linda Ramstedt. 2012.
\newblock Tapas: A multi-agent-based model for simulation of transport chains.
\newblock \emph{Simulation Modelling Practice and Theory}, 23:1--18.

\bibitem[{Huang et~al.(2023)Huang, Zhou, Chan, Fung, Wang, Zhang, Chang, and Ji}]{huang2023lvlms}
Kung-Hsiang Huang, Mingyang Zhou, Hou~Pong Chan, Yi~R Fung, Zhenhailong Wang, Lingyu Zhang, Shih-Fu Chang, and Heng Ji. 2023.
\newblock Do lvlms understand charts? analyzing and correcting factual errors in chart captioning.
\newblock \emph{arXiv preprint arXiv:2312.10160}.

\bibitem[{Jia et~al.(2021)Jia, Yang, Xia, Chen, Parekh, Pham, Le, Sung, Li, and Duerig}]{jia2021scaling}
Chao Jia, Yinfei Yang, Ye~Xia, Yi-Ting Chen, Zarana Parekh, Hieu Pham, Quoc Le, Yun-Hsuan Sung, Zhen Li, and Tom Duerig. 2021.
\newblock Scaling up visual and vision-language representation learning with noisy text supervision.
\newblock In \emph{International conference on machine learning}, pages 4904--4916. PMLR.

\bibitem[{Kafle et~al.(2018)Kafle, Price, Cohen, and Kanan}]{kafle2018dvqa}
Kushal Kafle, Brian Price, Scott Cohen, and Christopher Kanan. 2018.
\newblock Dvqa: Understanding data visualizations via question answering.
\newblock In \emph{Proceedings of the IEEE conference on computer vision and pattern recognition}, pages 5648--5656.

\bibitem[{Kantharaj et~al.(2022{\natexlab{a}})Kantharaj, Do, Leong, Tan, Hoque, and Joty}]{kantharaj2022opencqa}
Shankar Kantharaj, Xuan~Long Do, Rixie Tiffany~Ko Leong, Jia~Qing Tan, Enamul Hoque, and Shafiq Joty. 2022{\natexlab{a}}.
\newblock Opencqa: Open-ended question answering with charts.
\newblock \emph{arXiv preprint arXiv:2210.06628}.

\bibitem[{Kantharaj et~al.(2022{\natexlab{b}})Kantharaj, Leong, Lin, Masry, Thakkar, Hoque, and Joty}]{kantharaj2022chart}
Shankar Kantharaj, Rixie Tiffany~Ko Leong, Xiang Lin, Ahmed Masry, Megh Thakkar, Enamul Hoque, and Shafiq Joty. 2022{\natexlab{b}}.
\newblock Chart-to-text: A large-scale benchmark for chart summarization.
\newblock \emph{arXiv preprint arXiv:2203.06486}.

\bibitem[{Kim et~al.(2024)Kim, Yoon, In, Moon, Kim, and Park}]{kim2024adaptive}
Kibum Kim, Kanghoon Yoon, Yeonjun In, Jinyoung Moon, Donghyun Kim, and Chanyoung Park. 2024.
\newblock Adaptive self-training framework for fine-grained scene graph generation.
\newblock \emph{arXiv preprint arXiv:2401.09786}.

\bibitem[{Kim et~al.(2023)Kim, Yoon, Jeon, In, Moon, Kim, and Park}]{kim2023llm4sgg}
Kibum Kim, Kanghoon Yoon, Jaehyeong Jeon, Yeonjun In, Jinyoung Moon, Donghyun Kim, and Chanyoung Park. 2023.
\newblock Llm4sgg: Large language model for weakly supervised scene graph generation.
\newblock \emph{arXiv e-prints}, pages arXiv--2310.

\bibitem[{Lee et~al.(2023)Lee, Joshi, Turc, Hu, Liu, Eisenschlos, Khandelwal, Shaw, Chang, and Toutanova}]{lee2023pix2struct}
Kenton Lee, Mandar Joshi, Iulia~Raluca Turc, Hexiang Hu, Fangyu Liu, Julian~Martin Eisenschlos, Urvashi Khandelwal, Peter Shaw, Ming-Wei Chang, and Kristina Toutanova. 2023.
\newblock Pix2struct: Screenshot parsing as pretraining for visual language understanding.
\newblock In \emph{International Conference on Machine Learning}, pages 18893--18912. PMLR.

\bibitem[{Li et~al.(2022)Li, Li, Xiong, and Hoi}]{li2022blip}
Junnan Li, Dongxu Li, Caiming Xiong, and Steven Hoi. 2022.
\newblock Blip: Bootstrapping language-image pre-training for unified vision-language understanding and generation.
\newblock In \emph{International Conference on Machine Learning}, pages 12888--12900. PMLR.

\bibitem[{Li et~al.(2020)Li, Yin, Li, Zhang, Hu, Zhang, Wang, Hu, Dong, Wei et~al.}]{li2020oscar}
Xiujun Li, Xi~Yin, Chunyuan Li, Pengchuan Zhang, Xiaowei Hu, Lei Zhang, Lijuan Wang, Houdong Hu, Li~Dong, Furu Wei, et~al. 2020.
\newblock Oscar: Object-semantics aligned pre-training for vision-language tasks.
\newblock In \emph{Computer Vision--ECCV 2020: 16th European Conference, Glasgow, UK, August 23--28, 2020, Proceedings, Part XXX 16}, pages 121--137. Springer.

\bibitem[{Liu et~al.(2022{\natexlab{a}})Liu, Eisenschlos, Piccinno, Krichene, Pang, Lee, Joshi, Chen, Collier, and Altun}]{liu2022deplot}
Fangyu Liu, Julian~Martin Eisenschlos, Francesco Piccinno, Syrine Krichene, Chenxi Pang, Kenton Lee, Mandar Joshi, Wenhu Chen, Nigel Collier, and Yasemin Altun. 2022{\natexlab{a}}.
\newblock Deplot: One-shot visual language reasoning by plot-to-table translation.
\newblock \emph{arXiv preprint arXiv:2212.10505}.

\bibitem[{Liu et~al.(2022{\natexlab{b}})Liu, Piccinno, Krichene, Pang, Lee, Joshi, Altun, Collier, and Eisenschlos}]{liu2022matcha}
Fangyu Liu, Francesco Piccinno, Syrine Krichene, Chenxi Pang, Kenton Lee, Mandar Joshi, Yasemin Altun, Nigel Collier, and Julian~Martin Eisenschlos. 2022{\natexlab{b}}.
\newblock Matcha: Enhancing visual language pretraining with math reasoning and chart derendering.
\newblock \emph{arXiv preprint arXiv:2212.09662}.

\bibitem[{Liu et~al.(2023)Liu, Li, Wu, and Lee}]{liu2023visual}
Haotian Liu, Chunyuan Li, Qingyang Wu, and Yong~Jae Lee. 2023.
\newblock Visual instruction tuning.
\newblock \emph{arXiv preprint arXiv:2304.08485}.

\bibitem[{Lu et~al.(2019)Lu, Batra, Parikh, and Lee}]{lu2019vilbert}
Jiasen Lu, Dhruv Batra, Devi Parikh, and Stefan Lee. 2019.
\newblock Vilbert: Pretraining task-agnostic visiolinguistic representations for vision-and-language tasks.
\newblock \emph{Advances in neural information processing systems}, 32.

\bibitem[{Lu et~al.(2023{\natexlab{a}})Lu, Bansal, Xia, Liu, Li, Hajishirzi, Cheng, Chang, Galley, and Gao}]{lu2023mathvista}
Pan Lu, Hritik Bansal, Tony Xia, Jiacheng Liu, Chunyuan Li, Hannaneh Hajishirzi, Hao Cheng, Kai-Wei Chang, Michel Galley, and Jianfeng Gao. 2023{\natexlab{a}}.
\newblock Mathvista: Evaluating math reasoning in visual contexts with gpt-4v, bard, and other large multimodal models.
\newblock \emph{arXiv e-prints}, pages arXiv--2310.

\bibitem[{Lu et~al.(2023{\natexlab{b}})Lu, Peng, Cheng, Galley, Chang, Wu, Zhu, and Gao}]{lu2023chameleon}
Pan Lu, Baolin Peng, Hao Cheng, Michel Galley, Kai-Wei Chang, Ying~Nian Wu, Song-Chun Zhu, and Jianfeng Gao. 2023{\natexlab{b}}.
\newblock Chameleon: Plug-and-play compositional reasoning with large language models.
\newblock \emph{arXiv preprint arXiv:2304.09842}.

\bibitem[{Luo et~al.(2021)Luo, Li, Wang, and Lin}]{luo2021chartocr}
Junyu Luo, Zekun Li, Jinpeng Wang, and Chin-Yew Lin. 2021.
\newblock Chartocr: Data extraction from charts images via a deep hybrid framework.
\newblock In \emph{Proceedings of the IEEE/CVF winter conference on applications of computer vision}, pages 1917--1925.

\bibitem[{Masry et~al.(2023)Masry, Kavehzadeh, Do, Hoque, and Joty}]{masry2023unichart}
Ahmed Masry, Parsa Kavehzadeh, Xuan~Long Do, Enamul Hoque, and Shafiq Joty. 2023.
\newblock Unichart: A universal vision-language pretrained model for chart comprehension and reasoning.
\newblock \emph{arXiv preprint arXiv:2305.14761}.

\bibitem[{Masry et~al.(2022)Masry, Long, Tan, Joty, and Hoque}]{masry2022chartqa}
Ahmed Masry, Do~Xuan Long, Jia~Qing Tan, Shafiq Joty, and Enamul Hoque. 2022.
\newblock Chartqa: A benchmark for question answering about charts with visual and logical reasoning.
\newblock \emph{arXiv preprint arXiv:2203.10244}.

\bibitem[{Masry et~al.(2024)Masry, Shahmohammadi, Parvez, Hoque, and Joty}]{masry2024chartinstruct}
Ahmed Masry, Mehrad Shahmohammadi, Md~Rizwan Parvez, Enamul Hoque, and Shafiq Joty. 2024.
\newblock Chartinstruct: Instruction tuning for chart comprehension and reasoning.
\newblock \emph{arXiv preprint arXiv:2403.09028}.

\bibitem[{Meng et~al.(2024)Meng, Shao, Lu, Gao, Zhang, Qiao, and Luo}]{meng2024chartassisstant}
Fanqing Meng, Wenqi Shao, Quanfeng Lu, Peng Gao, Kaipeng Zhang, Yu~Qiao, and Ping Luo. 2024.
\newblock Chartassisstant: A universal chart multimodal language model via chart-to-table pre-training and multitask instruction tuning.
\newblock \emph{arXiv preprint arXiv:2401.02384}.

\bibitem[{Methani et~al.(2020)Methani, Ganguly, Khapra, and Kumar}]{methani2020plotqa}
Nitesh Methani, Pritha Ganguly, Mitesh~M Khapra, and Pratyush Kumar. 2020.
\newblock Plotqa: Reasoning over scientific plots.
\newblock In \emph{Proceedings of the IEEE/CVF Winter Conference on Applications of Computer Vision}, pages 1527--1536.

\bibitem[{Papineni et~al.(2002)Papineni, Roukos, Ward, and Zhu}]{papineni2002bleu}
Kishore Papineni, Salim Roukos, Todd Ward, and Wei-Jing Zhu. 2002.
\newblock Bleu: a method for automatic evaluation of machine translation.
\newblock In \emph{Proceedings of the 40th annual meeting of the Association for Computational Linguistics}, pages 311--318.

\bibitem[{Poco and Heer(2017)}]{poco2017reverse}
Jorge Poco and Jeffrey Heer. 2017.
\newblock Reverse-engineering visualizations: Recovering visual encodings from chart images.
\newblock In \emph{Computer graphics forum}, volume~36, pages 353--363. Wiley Online Library.

\bibitem[{Radford et~al.(2021)Radford, Kim, Hallacy, Ramesh, Goh, Agarwal, Sastry, Askell, Mishkin, Clark et~al.}]{radford2021learning}
Alec Radford, Jong~Wook Kim, Chris Hallacy, Aditya Ramesh, Gabriel Goh, Sandhini Agarwal, Girish Sastry, Amanda Askell, Pamela Mishkin, Jack Clark, et~al. 2021.
\newblock Learning transferable visual models from natural language supervision.
\newblock In \emph{International conference on machine learning}, pages 8748--8763. PMLR.

\bibitem[{Rajpurkar et~al.(2016)Rajpurkar, Zhang, Lopyrev, and Liang}]{rajpurkar2016squad}
Pranav Rajpurkar, Jian Zhang, Konstantin Lopyrev, and Percy Liang. 2016.
\newblock Squad: 100,000+ questions for machine comprehension of text.
\newblock \emph{arXiv preprint arXiv:1606.05250}.

\bibitem[{Shao et~al.(2023)Shao, Yu, Wang, and Yu}]{shao2023prompting}
Zhenwei Shao, Zhou Yu, Meng Wang, and Jun Yu. 2023.
\newblock Prompting large language models with answer heuristics for knowledge-based visual question answering.
\newblock In \emph{Proceedings of the IEEE/CVF Conference on Computer Vision and Pattern Recognition}, pages 14974--14983.

\bibitem[{Su et~al.(2019)Su, Zhu, Cao, Li, Lu, Wei, and Dai}]{su2019vl}
Weijie Su, Xizhou Zhu, Yue Cao, Bin Li, Lewei Lu, Furu Wei, and Jifeng Dai. 2019.
\newblock Vl-bert: Pre-training of generic visual-linguistic representations.
\newblock \emph{arXiv preprint arXiv:1908.08530}.

\bibitem[{Tan and Bansal(2019)}]{tan2019lxmert}
Hao Tan and Mohit Bansal. 2019.
\newblock Lxmert: Learning cross-modality encoder representations from transformers.
\newblock \emph{arXiv preprint arXiv:1908.07490}.

\bibitem[{Wei et~al.(2022)Wei, Wang, Schuurmans, Bosma, Xia, Chi, Le, Zhou et~al.}]{wei2022chain}
Jason Wei, Xuezhi Wang, Dale Schuurmans, Maarten Bosma, Fei Xia, Ed~Chi, Quoc~V Le, Denny Zhou, et~al. 2022.
\newblock Chain-of-thought prompting elicits reasoning in large language models.
\newblock \emph{Advances in Neural Information Processing Systems}, 35:24824--24837.

\bibitem[{Xia et~al.(2023)Xia, Zhang, Peng, Liao, Ye, Shi, Yan, and Qiao}]{xia2023structchart}
Renqiu Xia, Bo~Zhang, Haoyang Peng, Ning Liao, Peng Ye, Botian Shi, Junchi Yan, and Yu~Qiao. 2023.
\newblock Structchart: Perception, structuring, reasoning for visual chart understanding.
\newblock \emph{arXiv preprint arXiv:2309.11268}.

\bibitem[{Xue et~al.(2023)Xue, Chen, Yu, Chen, Zhou, and Peng}]{xue2023chartdetr}
Wenyuan Xue, Dapeng Chen, Baosheng Yu, Yifei Chen, Sai Zhou, and Wei Peng. 2023.
\newblock Chartdetr: A multi-shape detection network for visual chart recognition.
\newblock \emph{arXiv preprint arXiv:2308.07743}.

\bibitem[{Zhang et~al.(2021)Zhang, Li, Hu, Yang, Zhang, Wang, Choi, and Gao}]{zhang2021vinvl}
Pengchuan Zhang, Xiujun Li, Xiaowei Hu, Jianwei Yang, Lei Zhang, Lijuan Wang, Yejin Choi, and Jianfeng Gao. 2021.
\newblock Vinvl: Revisiting visual representations in vision-language models.
\newblock In \emph{Proceedings of the IEEE/CVF conference on computer vision and pattern recognition}, pages 5579--5588.

\bibitem[{Zhou et~al.(2020)Zhou, Palangi, Zhang, Hu, Corso, and Gao}]{zhou2020unified}
Luowei Zhou, Hamid Palangi, Lei Zhang, Houdong Hu, Jason Corso, and Jianfeng Gao. 2020.
\newblock Unified vision-language pre-training for image captioning and vqa.
\newblock In \emph{Proceedings of the AAAI conference on artificial intelligence}, volume~34, pages 13041--13049.

\bibitem[{Zhou et~al.(2023)Zhou, Fung, Chen, Thomas, Ji, and Chang}]{zhou2023enhanced}
Mingyang Zhou, Yi~R Fung, Long Chen, Christopher Thomas, Heng Ji, and Shih-Fu Chang. 2023.
\newblock Enhanced chart understanding in vision and language task via cross-modal pre-training on plot table pairs.
\newblock \emph{arXiv preprint arXiv:2305.18641}.

\bibitem[{Zhu et~al.(2023)Zhu, Chen, Shen, Li, and Elhoseiny}]{zhu2023minigpt}
Deyao Zhu, Jun Chen, Xiaoqian Shen, Xiang Li, and Mohamed Elhoseiny. 2023.
\newblock Minigpt-4: Enhancing vision-language understanding with advanced large language models.
\newblock \emph{arXiv preprint arXiv:2304.10592}.

\bibitem[{Zhu et~al.(2024)Zhu, Jia, Zhang, Li, and Jiang}]{zhu2024multichartqa}
Zifeng Zhu, Mengzhao Jia, Zhihan Zhang, Lang Li, and Meng Jiang. 2024.
\newblock Multichartqa: Benchmarking vision-language models on multi-chart problems.
\newblock \emph{arXiv preprint arXiv:2410.14179}.

\end{thebibliography}

\newpage

\appendix

\section{Complete Related Work}
\label{app:relatedwork}

\noindent\textbf{Pre-trained vision-language models.} 
Vision-language pre-training seeks to enhance the performance of downstream vision and language tasks such as visual question answering (VQA) \citep{tan2019lxmert, alayrac2022flamingo}, image-text retrieval \citep{chen2020uniter, jia2021scaling}, and image captioning \citep{li2020oscar, li2022blip}.

Several models learn cross-modal representation from visual features and language tokens \citep{chen2020uniter, su2019vl}. Specifically, \citet{lu2019vilbert} employs two single-modal networks, alongside a cross-modal transformer layer integrating information from input sentences and images.

Others propose a different approach to develop an enhanced Large Multimodal Model (LMM). For example, \citet{liu2023visual} employs a visual encoder and a language decoder from other studies \citep{radford2021learning, chiang2023vicuna}. It is capable of performing various QA tasks including conversation, detail description, complex reasoning and focuses on chat capability.

\smallskip
\noindent\textbf{Chart reasoning models.} 
Early methods for processing charts predominantly utilized rule-based approaches \citep{balaji2018chart, choi2019visualizing}. However, they were specifically tailored to predefined chart formats, limiting their applicability only to certain types of charts.
Moreover, incorporating new chart designs require additional rules, leading to the impracticality of immediate adaptation to new chart formats.

Due to above limitations, researchers utilize modules such as OCR and key-point detection \citep{poco2017reverse, zhou2023enhanced, xue2023chartdetr}. However, these methods can be time-consuming and require dataset annotation for labeling, and rely on the performance of modules. 
For example, \citet{luo2021chartocr} suffers from a drawback as it solely predicts the raw data values without establishing connections to their respective axes or legends.

In an effort to address these challenges, researchers develop vision-language models \citep{meng2024chartassisstant, huang2023lvlms} without heuristic rules. Some approaches comprehend the chart and respond to questions in natural language \citep{masry2023unichart, han2023chartllama}. However, such models require fine-tuning for each downstream task, constraining their adaptability for diverse tasks. 
Meanwhile, studies such as \citet{methani2020plotqa, lu2023mathvista} have created datasets to enable more realistic mathematical reasoning.

To tackle these challenges, recent studies  \citet{liu2022deplot, xia2023structchart} introduce a two-step approach, chart-to-table extraction and reasoning with LLM.
Recent advancements like \citet{liu2022deplot, xia2023structchart} have introduced a two-step approach to tackle challenges in chart analysis. Firstly, charts are transformed into tables, improving interpretability and aiding in identifying inaccuracies. 
\citet{lu2023chameleon} utilizes Program of Thought (PoT), achieving significant performance gains. This approach enables more accurate and interpretable reasoning compared to traditional methods, indicating that table extraction is crucial.

\section{Row-column Rendering}
\label{app:rendering}

\begin{figure*}[!t]  %%% t: top, b: bottom, h: here
\centering
    \includegraphics[width=0.95\linewidth]{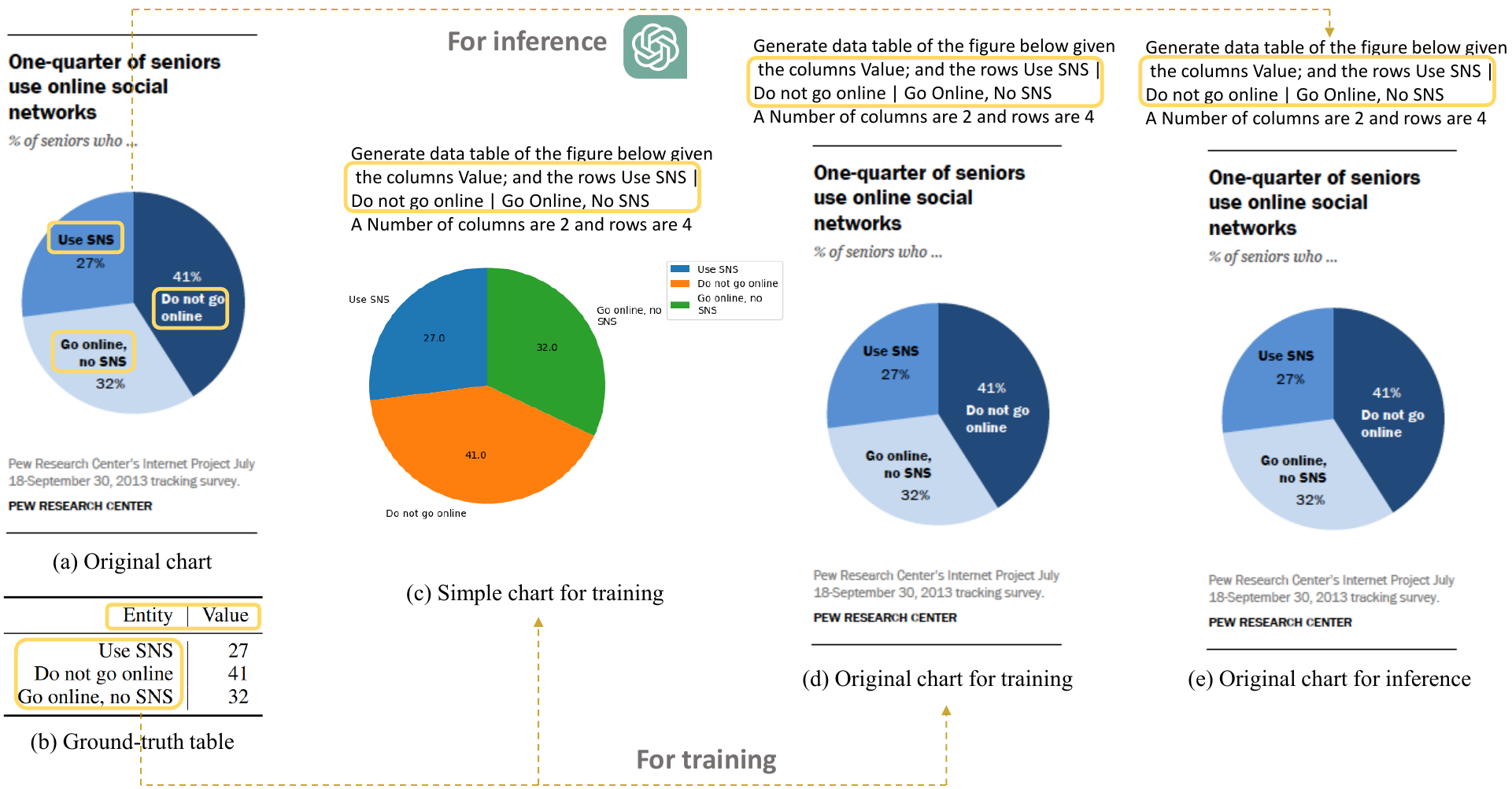} 
    \caption{An example of row-column rendering for an input image in training and inference stage.}
    \label{fig:rendering}
    %{-2ex}
\end{figure*}

Fig.~\ref{fig:rendering} presents a detailed example of our row-column rendering process. For a simple chart and an original chart used in the training stage, we extract rows and columns from their associated table and directly render them onto the charts since the ground-truth table is provided in this stage. However, for the original chart used in the inference stage, since we cannot access the ground-truth table, we use an LMM to extract rows and columns from the input chart and rendered onto the image.

\begin{figure*}[!h]  %%% t: top, b: bottom, h: here
\centering
    \includegraphics[width=0.95\linewidth]{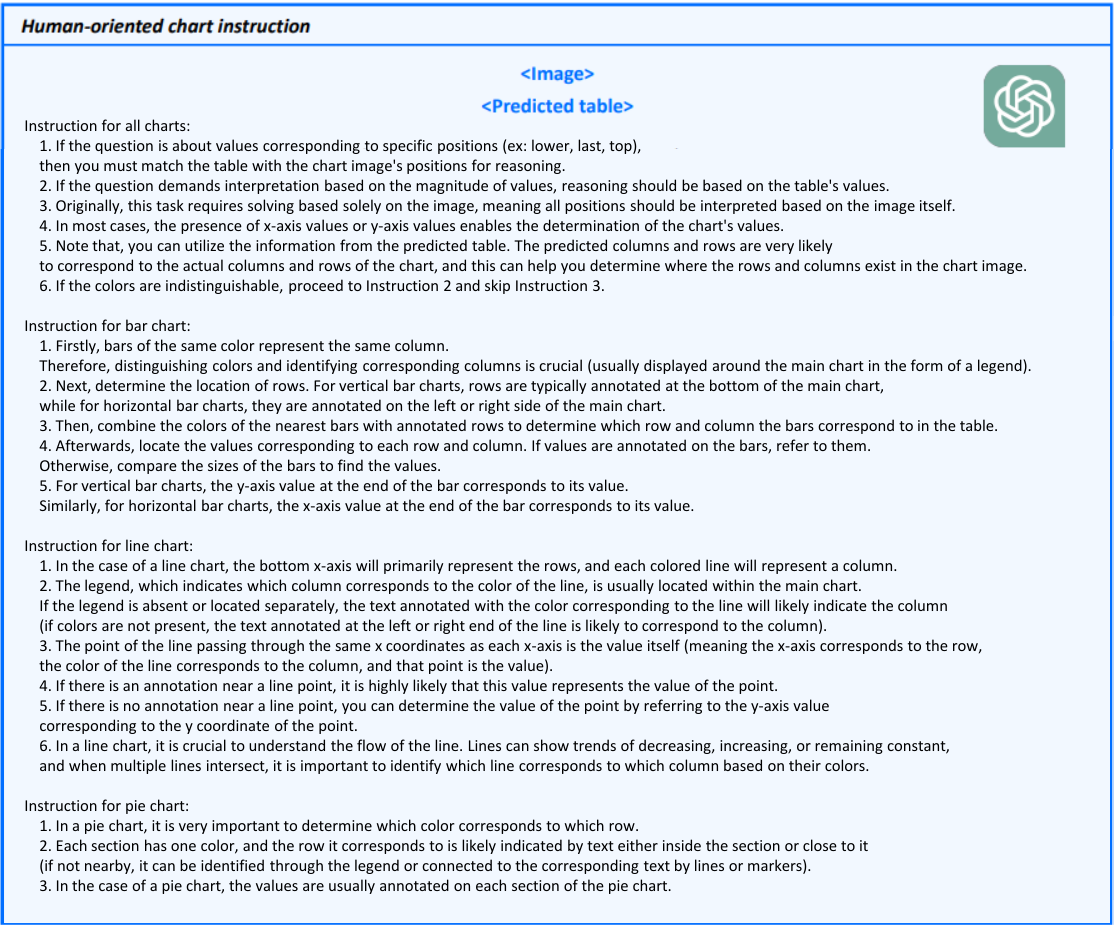} 
    \caption{Overall prompt of \prompt.}
    \label{fig:prompt}
    %{-2ex}
\end{figure*}

\section{Human-oriented Prompt for Chart Reasoning}
\label{app:prompt}

We present instructions in the prompt to mimic the way humans commonly perceive charts. We name this approach \prompt, which is a simple yet effective methodology as shown in the QA results. The detailed prompt is shown in Fig.~\ref{fig:prompt}. 

In the \prompt, universal instructions applicable to all chart types are provided first. They include components contained in every chart such as chart title, y-axis, x-axis, legend, and data labels, facilitating the LMM to comprehend the chart. Furthermore, we guide the alignment between a chart and the generated table by instructing the system within the prompt to effectively comprehend the positions of columns and rows as well as the outline of the chart, referring to the table generated by \proposed.

Subsequently, instructions are described for each chart type. Here, we describe the instructions designed for bar charts, but those for line charts and pie charts are designed in a similar manner. 

\smallskip
\noindent\textbf{Bar Chart: Instruction 1.}
When humans recognize a bar chart, the initial observation would be columns distinguished by their colors. Following this observation, the legend within the image is consulted to identify which color denotes which column. If the colors are indistinguishable, proceed to Instruction 2 and skip Instruction 3.

\smallskip
\noindent\textbf{Bar Chart: Instruction 2.}
Having identified the columns, humans would proceed to observe which row each bar corresponds to. Specifically, for vertical bar graphs, rows are usually annotated below the main chart, whereas for horizontal bar graphs, annotations on the lateral sides determine the rows.

\smallskip
\noindent\textbf{Bar Chart: Instruction 3.}
Humans then would match the color of the bar with the annotated row, thereby aligning each bar with its corresponding column and row.

\smallskip
\noindent\textbf{Bar Chart: Instruction 4 \& 5.}
Lastly, humans would figure out the value corresponding to each row and column by referencing the value annotated on the bar, the relative sizes between bars, and the x-axis and y-axis. Notably, it is understood intuitively that the value at the point where the bar terminates on the x-axis or y-axis represents the value of that bar. 

We argue that developing a language model that captures the sequential and unconscious reasoning process involved in human chart recognition through \prompt~not only mimics human cognition effectively but also ensures the alignment of predicted tables with chart images. This leads to substantial performance enhancements at a lower cost.

\section{Details regarding Experiments}
\label{app:setting}

\subsection{Dataset Description}\label{app:dataset}

In this paper, we mainly evaluate our proposed method with a widely-used dataset, ChartQA \citep{masry2022chartqa}. It consists of real-world charts containing complex information, which aligns well with our motivation compared to other synthetic datasets such as PlotQA \citep{methani2020plotqa} and DVQA \citep{kafle2018dvqa}, both of which only include simple charts. Furthermore, ChartQA is comprised of three chart types (i.e., pie, bar, and line), whereas DVQA and PlotQA comprise only one or two types of chart. Therefore, ChartQA is a more challenging dataset and is desirable for evaluating performance on real-world chart reasoning tasks. 

For these reasons, we mainly focus on ChartQA dataset in this paper since other datasets are composed only of very simple charts, which are not suitable for our model that is designed to handle real-world charts containing noisy information unnecessary for reasoning. Nevertheless, to demonstrate that \proposed~can be applied to any dataset, we conduct additional experiments on the PlotQA dataset in the Appendix \ref{app:plotqaexperiment}.

ChartQA also includes human-authored and LLM-augmented QA pairs for reasoning. Please note that we do not use QA pairs for training, but train the model to generate tables only from the charts in the training set, then use QA pairs from the test set for measuring reasoning performance. 

Detailed statistics of ChartQA dataset are presented in Table \ref{tab:dataset}, and statistics of PlotQA dataset are presented in Table \ref{tab:datasetplotqa}. To reduce the training cost and balance the dataset with ChartQA, we stratified and sampled 10\% of the images from the PlotQA dataset based on type, which were then used for training and inference. Additionally, for QA pairs, we used one pair per image.

\begin{table}[h]
\caption{Statistics of ChartQA dataset.}
\label{tab:dataset}
\small
\resizebox{1\linewidth}{!}{
\centering
\begin{tabular}{c|ccc|c}
\toprule
\diagbox{split}{type} & Pie& Bar  & Line & QA pair \\
\midrule
Train set & 541 & 15,581 & 2,195 & - \\
Validation set & 48 & 837 & 171 & - \\
Test set& 78 & 1,230 & 211 & 2,500 \\
\bottomrule
\end{tabular}
}
\end{table}

\begin{table}[h]
\caption{Statistics of PlotQA dataset.}
\label{tab:datasetplotqa}
\small
\resizebox{1\linewidth}{!}{
\centering
\begin{tabular}{c|ccc|c}
\toprule
\diagbox{split}{type} & Dot line & Line  & Bar & QA pair \\
\midrule
Train set & 26,010 & 25,897 & 105,163 & - \\
Validation set & 5,571 & 5,547 & 22,541 & - \\
Test set & 5,574 & 5,549 & 22,534 & 4,342,514 \\
\bottomrule
\end{tabular}
}
\end{table}

\subsection{Compared Methods}\label{app:baseline}
We use a wide range of models capable of performing chart reasoning as baselines, categorizing them into three distinct categories. 

The first category includes Vision-Language Pre-trained models (VLP) as follows:
\begin{itemize}
    %{-1ex}
    \item TaPas \citep{holmgren2012tapas}
    %{-1ex}
    \item V-TaPas \citep{masry2022chartqa}
    %{-1ex}
    \item T5 \citep{bujard198726}
    %{-1ex}
    \item VL-T5 \citep{cho2021unifying}
    %{-1ex}
    \item PaLI \citep{chen2022pali}
    %{-1ex}
    \item Mini-GPT \citep{zhu2023minigpt}
    %{-1ex}
    \item LLaVa \citep{liu2023visual}
    %{-1ex}
    \item GPT-4V \citep{achiam2023gpt}
    %{-1ex}
\end{itemize}

The second category consists of supervised models, including followings:

\begin{itemize}
%{-1ex}
    \item ChartQA \citep{masry2022chartqa}
    %{-1ex}
    \item ChartT5 \citep{zhou2023enhanced}
    %{-1ex}
    \item Pix2Struct \citep{lee2023pix2struct}
    %{-1ex}
    \item MatCha \citep{liu2022matcha}
    %{-1ex}
    \item Unichart \citep{masry2023unichart}
    %{-1ex}
    \item ChartLlama \citep{han2023chartllama}
    \item ChartAssistant \citep{meng2024chartassisstant}
    \item ChartInstruct \citep{masry2024chartinstruct}
\end{itemize}

\begin{table}[h]
  \centering
  \resizebox{0.9\columnwidth}{!}{
  \begin{tabular}{lccc}
    \toprule
    & LLM & Textual info. & Designed prompt \\
    \midrule
    Unichart & \ding{55} & \ding{55} & \ding{55} \\
    Deplot & \ding{51} & \ding{55} & \ding{55} \\
    \proposed & \ding{51} & \ding{51} & \ding{51} \\
    \bottomrule
  \end{tabular}
  }
  \caption{Comparison of models based on their utilization of LLM, textual information, and designed prompt.}
  \label{tab:modelcomparison}
\end{table}

The third category consists of table-based reasoning models, including Deplot \citep{liu2022deplot} and Unichart \citep{masry2023unichart}. Deplot converts charts to tables similar to \proposed~and employs prompt-based methodologies such as Chain-of-Thought (CoT) \citep{wei2022chain}, Program-of-Thought (PoT) \citep{chen2022program} alongside using an LLM for question reasoning. Although Unichart is trained to perform QA directly from charts, it is also pre-trained to generate tables. In this paper, we use this variant of Unichart to extract tables and utilize LLM to fairly compare with Deplot in the same setting.

Regardless of our application of Unichart in the same setting as Deplot, the original Unichart requires direct training for each downstream task. It has limitations in terms of adaptability and may exhibit suboptimal performance. Deplot and \proposed~achieve performance gains and enhanced adaptability by performing inference through an LLM after chart-to-table extraction. Additionally, \proposed~significantly improves the accuracy of table extraction and downstream tasks by leveraging textual information and utilizing prompts specifically designed for charts.
The differences among the aforementioned models can be referred to Table \ref{tab:modelcomparison}.

As the existing chart reasoning models \citep{liu2022deplot, masry2023unichart, masry2024chartinstruct, meng2024chartassisstant, liu2022matcha, masry2022chartqa, lee2023pix2struct} require expensive computational time, they have reported the performance after a single run, which makes it difficult to conduct statistical comparison. Following the previous works, we only report the performance after a single run. However, we claim that our results improve the performance of chart question answering with significant margin.

\subsection{Details regarding Proposed Metric}\label{app:metric}

\begin{figure}[!t]  %%% t: top, b: bottom, h: here
\centering
    \includegraphics[width=0.99\linewidth]{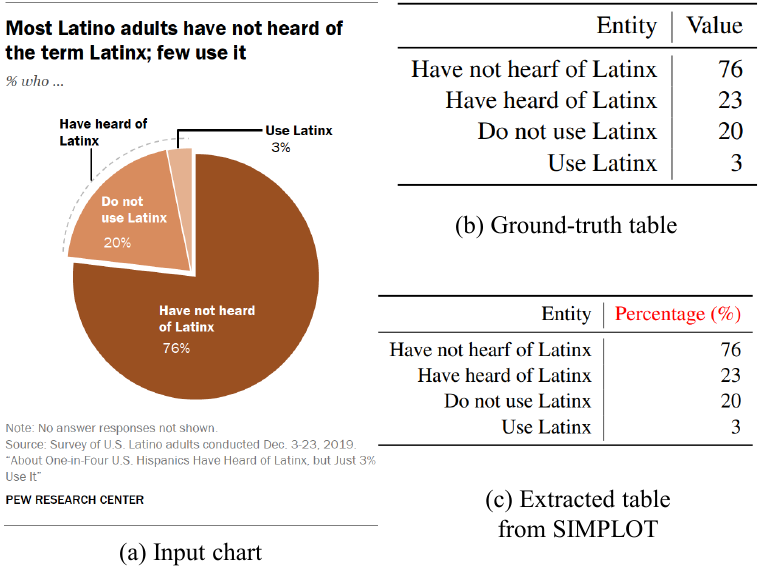} 
    %{-3ex}
    \caption{An example of an extracted table that poses no issues for reasoning despite having a different column name from that in the ground-truth table.}
    \label{fig:rd1}
    %{-1ex}
\end{figure}

\begin{figure}[!t]  %%% t: top, b: bottom, h: here
\centering
    \includegraphics[width=0.8\linewidth]{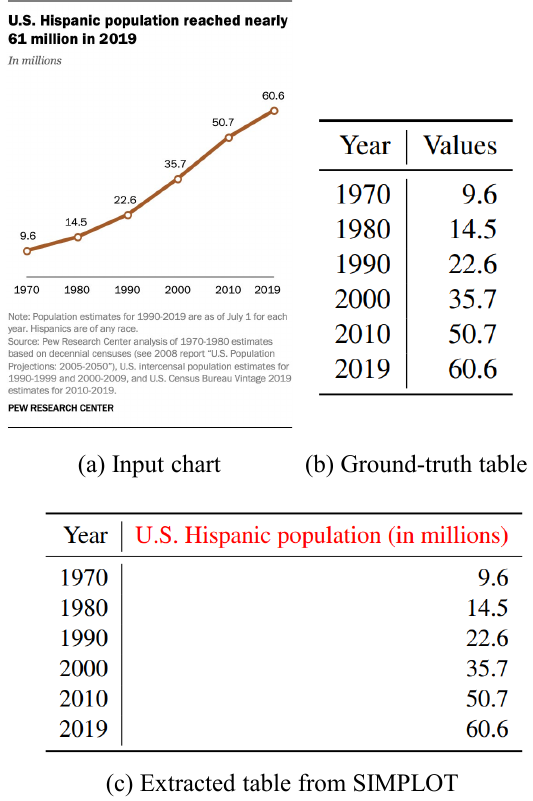} 
    \caption{An example where reasoning remains unaffected, yet a column extracted by \proposed~differs significantly due to the lack of clear column description in the image.}
    \label{fig:rd2}
    %{-2ex}
\end{figure}

As illustrated in Fig. \ref{fig:rd1}, columns of extracted table from \proposed~(e.g., "Percentage (\%)" in the figure) often carry the same meaning  as the ground truth (e.g., "Value" in the figure) but have different Levenshtein distances. Additionally, as shown in Fig.\ref{fig:rd2}, in cases where columns are not explicitly stated in the image, an \proposed~utilizing LMM without further fine-tuning generates arbitrary columns. We confidently assert that there are unequivocally no problems associated with chart analysis when employing an LMM in both scenarios. However, given that $RMS$ is composed of the product of Levenshtein distance (from columns and rows) and relative distance (from values), this leads to an apparent degradation in performance of chart-to-table task even when there are no issues with QA. Consequently, we construct a Levenshtein distance matrix by concatenating column and row predictions with the ground truth, followed by applying minimal cost matching to consider only the relative distance of each pair. We adopt this Relative Distance ($RD$) as our main metric. This approach allows us to accurately assess performance in value mapping.

We follow Deplot \citep{liu2022deplot} to define RD as in the following equation, with most formulas extracted from this paper. Following \citet{liu2022deplot}, we use basic concept of Related Mapping Similarity ($RMS$), regarding tables as unordered sets of mappings from row and column headers ($r,c$) to a single value $v$. $p_i=(p^r_i,p^c_i,p^v_i)$ and $t_j=(t^r_j,t^c_j,t^v_j)$ indicates each entity for the predicted table $\mathcal{P}=\{p_i\}_{1\leq i \leq N}$ and the ground truth table $\mathcal{T}=\{t_j\}_{1\leq j \leq M}$, respectively. Utilizing \textit{Normalized Levenshtein Distance} (${NL}_{\tau}$) \citep{biten2019icdar}, we denote the distance between two keys $p_i$ and $t_j$ as ${NL}_{\tau}(p^r||p^c,t^r||t^c)$ where $||$ indicates concatenation of string. The distance between values of table is computed with relative distance $D_{\theta}(p^v,t^v)=min(1,||p^v-t^v||/||t^v||)$ and distances larger than $\theta$ are set to the maximum of 1. Incorporating these two distances, we obtain the similarity between two entities in a mapping $D_{\tau,\theta}(p,t)$ as $(1-{NL}_{\tau}(p^r||p^c,t^r||t^c))(1-D_{\theta}(p^v,t^v))$. 

For $RMS$, \citet{liu2022deplot} computes the pairwise similarity in $\mathcal{P}$ and $\mathcal{T}$ using the cost function $(1-{NL}_{\tau}(p^r||p^c,t^r||t^c))$, obtaining a similarity matrix in shape of $N\times M$ and identifying the minimal cost matching $\mathbf{X} \in \mathbb{R}^{N \times M}$ between the keys (in the binary matrix form).

While precision and recall is obtained from two mappings in Deplot \cite{liu2022deplot}, we only use distance of the numeric entries to compute $RD$:

\begin{equation}\label{eq:rd_precision}
\small
    RD_{precision} = 1 - \frac{\sum_{i=1}^{N}\sum_{j=1}^{M} \mathbf{X}_{ij}(D_{\theta}(p^v,t^v))}{N},
\end{equation}

\begin{equation}\label{eq:rd_recall}
\small
    RD_{recall} = 1 - \frac{\sum_{i=1}^{N}\sum_{j=1}^{M} \mathbf{X}_{ij}(D_{\theta}(p^v,t^v))}{M}.
\end{equation}

Similar to Deplot, we compute ${RD}_{F1}$ as the harmonic mean of the precision and recall.

\section{Additional Experiments with PlotQA}\label{app:plotqaexperiment}

To demonstrate that our method can be applied to a variety of data beyond the ChartQA dataset, we present additional experimental results on the PlotQA dataset in Table \ref{tab:plotqaperformance}. Our proposed method exhibits superior performance across various chart types, indicating that it can be applied to any chart type. However, unlike ChartQA, which consists of real-world charts, PlotQA is composed only of very simple charts, so the performance differences between \proposed~and other methods are not as significant as in the result on the ChartQA dataset. We emphasize once again that \proposed~is designed to handle real-world charts containing unnecessary and noisy information for chart question answering.

\begin{table}[h]
\caption{Chart question answering performance (RA) on the PlotQA dataset.}
\label{tab:plotqaperformance}
\small
% \resizebox{1\linewidth}{!}{
\centering
\begin{tabular}{l|cccc}
\toprule
Models & Dot line & Line & Bar & Overall  \\
\midrule
GPT-4V &  50.53 & \underline{58.84}  & 53.85  & 54.11 \\
Unichart & 58.78 & 53.26  & 60.10  & 58.74 \\
Deplot & \textbf{66.66}  & 55.59  & \underline{61.73}  & \underline{61.53}  \\
\textbf{\proposed} & \underline{60.93} &  \textbf{65.57} & \textbf{73.84}  &  \textbf{70.32} \\
\bottomrule
\end{tabular}
% }
\end{table}

\section{Effectiveness of Utilizing Image with Extracted Table}\label{app:limitation}

Using tables for chart reasoning allows for performing various downstream tasks (such as chart summarization, chartQA, etc.). However, there is an inherent limitation of using tables for chart reasoning, i.e., incorrect extraction of the table leads to failure in all subsequent downstream tasks. Taking this limitation into account, unlike Deplot that solely relies on tables for reasoning,~\proposed~simultaneously considers both the extracted table and the image itself to compensate for the incorrectly extracted table.

To further verify the effectiveness of using an image alongside the generated table, we conduct additional experiments. Table \ref{tab:effectofimage} proves that using images along with tables increases the success rate of QA compared to using only tables. We assume both Deplot and \proposed~failed to accurately extract the table in cases where the Relative Distance ($RD$) was below 0.7 (i.e., when the \proposed~does not benefit from utilizing simple charts) and the difference between the two RDs was within 0.1, and there are 155 such cases. Among these cases, even though we provide images to Deplot and removed the prompt from \proposed~for a fair comparison, while Deplot failed in QA, \proposed~succeeded in 48 cases, which is more than 30\% additional correct answers. Conversely, when \proposed~failed in QA, Deplot succeeded in only 7 cases, which is less than 5\%. This demonstrates that utilizing both plot and table can effectively address the limitation of solely relying on tables for reasoning.

\begin{table}[h]
\caption{The number of times QA succeeded or failed when both methods failed to extract the table properly.}
\label{tab:effectofimage}
\small
% \resizebox{1\linewidth}{!}{
\centering
\begin{tabular}{c|cc}
\toprule
\diagbox{Deplot}{\proposed} & Success & Fail  \\
\midrule
Success & - & 7 \\
Fail & 48 & - \\
\bottomrule
\end{tabular}
% }
\end{table}

\section{Hyperparameter Analysis}\label{app:lambda}

\begin{figure}[!t]  %%% t: top, b: bottom, h: here
\centering
    %{-1ex}
    \includegraphics[width=1\linewidth]{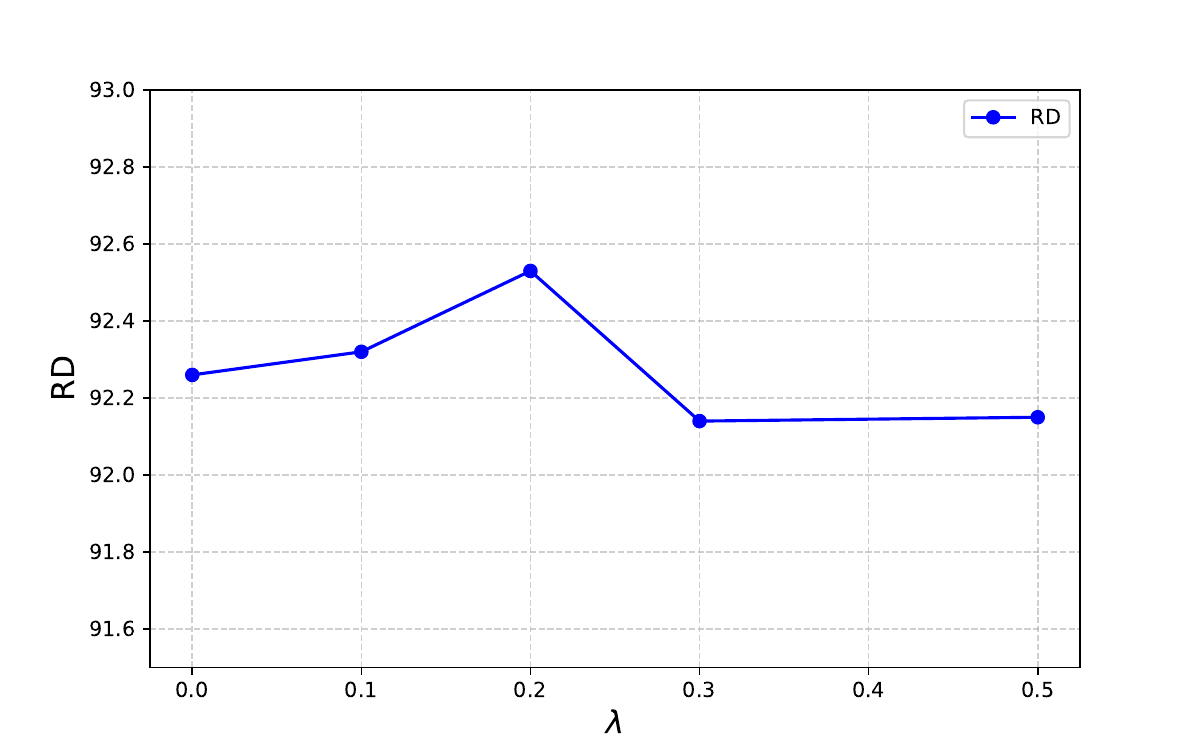} 
    %{-1ex}
    \caption{Table extraction performance ($RD_{F1}$) with varying $\lambda$.}
    \label{fig:rdlambda}
\end{figure}

We conduct a hyperparameter anaylsis by setting various values for the balancing parameter $\lambda$ and analyzing the result changes in RD. As illustrated in Fig. \ref{fig:rdlambda}, the changes in RD values due to variations in $\lambda$ are minimal. This indicated that our model is robust and performs consistently well regardless of the $\lambda$ value. 

\section{Effectiveness of Negative Sample}\label{app:negative}

For more effective learning, we can generate negative samples at no additional cost and use them for training. Charts of the same type (e.g., line charts) tend to have a similar format even if they contain different content, so the representations obtained from the model will likely be similar. Since the value information within the chart is critical, we utilize negative samples that have a format similar to positive samples but contain different values, allowing the model to capture finer differences. In other words, the purpose of utilizing negative samples is to more precisely discriminate between data that may be mapped closely in the representation space because the charts look similar. The effect of using negative samples together is shown in Table \ref{tab:negative}. 
% For example, the goal could be to reduce the error in the values by 1. 
Although the difference in table extraction performance may appear minor, even a slight improvement in extracting values more accurately is important in the process of converting charts into tables.

\begin{table}[h]
  \centering
  \resizebox{0.5\columnwidth}{!}{
  \begin{tabular}{cc|c}
    \toprule
     Positive & Negative & $RD_{F1}$ \\
     \midrule
     \ding{51} & \ding{55}  & 92.18 \\
     \ding{51} & \ding{51} & 92.32  \\
  \bottomrule
\end{tabular}
}
\caption{Ablation studies on utilizing negative sample ~\proposed~for chart-to-table extraction.}
\label{tab:negative}
%{-2.5ex}
\end{table}

\section{More Case Studies}\label{app:casestudy}

\begin{figure}[!t]  %%% t: top, b: bottom, h: here
\centering
    %{-1ex}
    \includegraphics[width=0.9\linewidth]{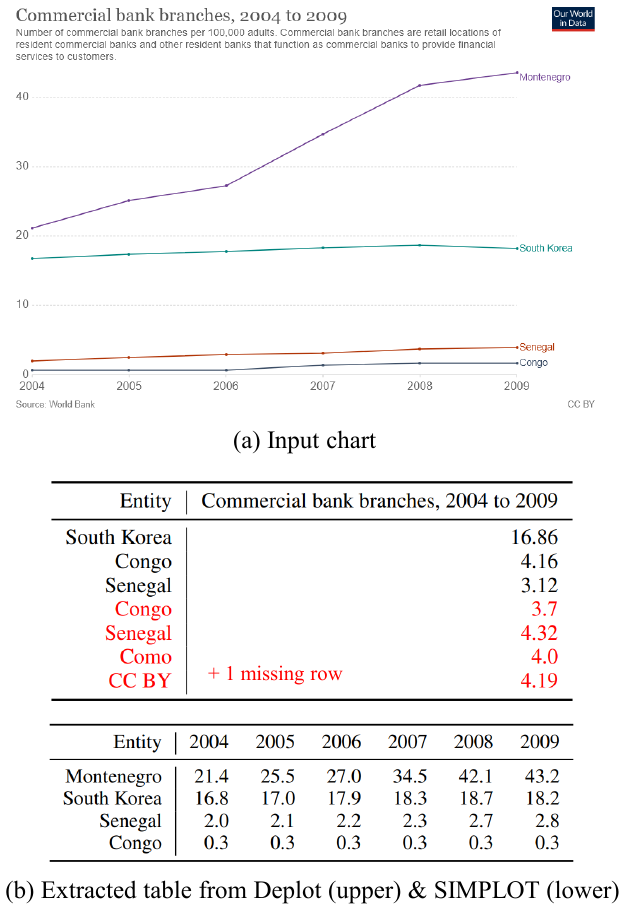} 
    %{-1ex}
    \caption{Case study. Extract precise rows without unnecessary information in the chart.}
    \label{fig:case2}
    %{-2ex}
\end{figure}

Here, we present more case studies. We confirm that rendering columns and rows of a table onto the image is beneficial for extracting an accurate table. In Fig. \ref{fig:case2}, Deplot \citep{liu2022deplot} omits a relevant row (i.e., "Montenegro"), redundantly extracts several rows, and even generates entirely irrelevant rows such as "CC BY," which would result in a poor performance in reasoning tasks.
In contrast, \proposed~accurately generates a table by extracting only the relevant rows, ensuring accurate reasoning.

\begin{figure}[h]  %%% t: top, b: bottom, h: here
\centering
    %{-1ex}
    \includegraphics[width=0.99\linewidth]{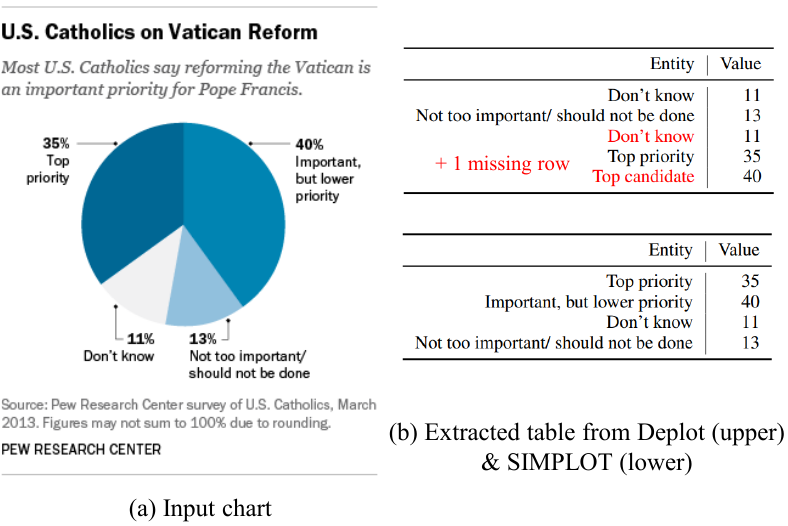} 
    %{-3ex}
    \caption{Case study. Extract all necessary rows in the pie chart.}
    \label{fig:case3}
    %{-ex}
\end{figure}

Fig.~\ref{fig:case3} and \ref{fig:case4} show that in both pie charts and bar charts, Deplot \citep{liu2022deplot} fails to capture relevant rows, generates inaccurate rows, thus failing to create values. On the other hand, \proposed~consistently produces accurate tables. This directly proves the generality of~\proposed~across various chart types.

\begin{figure*}[t]  %%% t: top, b: bottom, h: here
\centering
    %{-1ex}
    \includegraphics[width=0.99\linewidth]{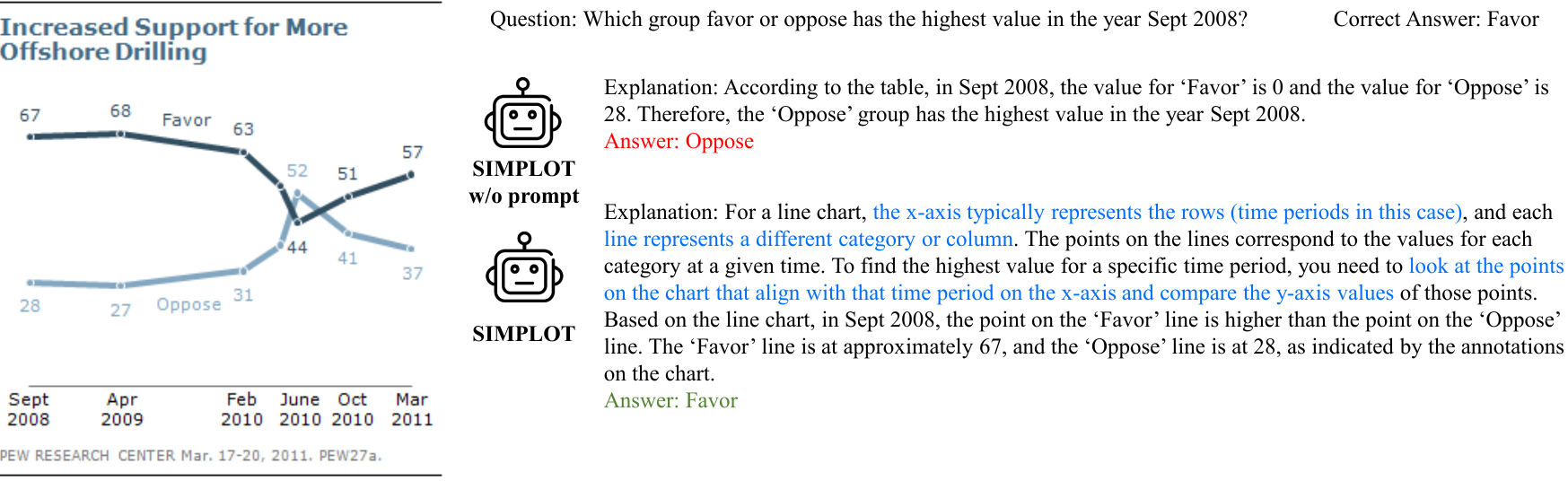} 
    %{-3ex}
    \caption{Case study. Interpreting charts similar to how humans do using \prompt.}
    \label{fig:case6}
    %{-2.5ex}
\end{figure*}

Furthermore, Fig.~\ref{fig:case6} demonstrates that our \prompt~works well by imitating the way of human interpretation for charts. Without the proposed prompt, an LLM fails to detect the exact location of the line component "Favor", leading to a wrong answer. However, utilizing \prompt, an LMM can precisely locate the necessary component, resulting in a correct answer.

\section{Evaluations on more Challenging Questions}\label{app:challenging}
\begin{figure}[h]  %%% t: top, b: bottom, h: here
\centering
    %{-1ex}
    \includegraphics[width=0.99\linewidth]{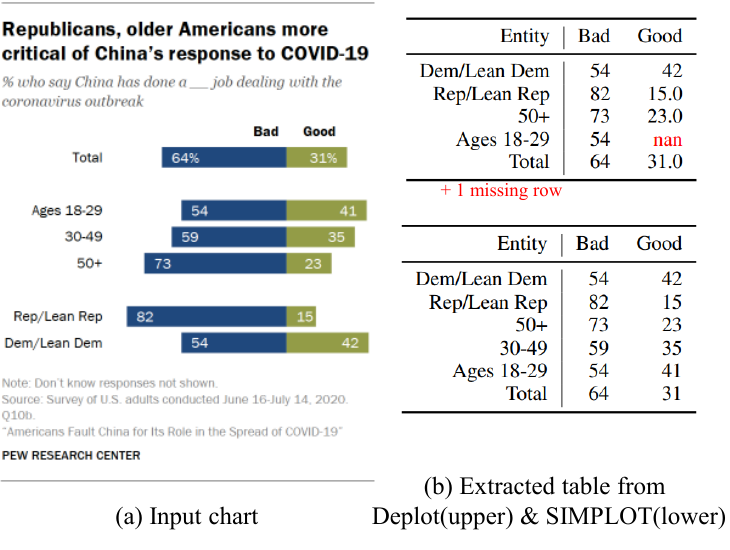} 
    %{-3ex}
    \caption{Case study. Extract all necessary rows and precise value in the bar chart.}
    \label{fig:case4}
    %{-2.5ex}
\end{figure}

While conducting case studies, we observed that Deplot and Simplot often exhibited similar QA performance, despite Deplot failing to accurately extract tables that \proposed~successfully extract. \textit{We attribute this mainly to the simplicity of the questions in the ChartQA dataset.}
For example, in Fig. \ref{fig:case4}, although Deplot fails to accurately extract the table, if the question happens to be about `Dem/Lean Dem', which is not related to the inaccurate part of the extracted table, Deplot can still generate the correct answer. This proves that the performance of our method can be further differentiated with more complex questions.

\begin{figure*}[!t]  %%% t: top, b: bottom, h: here
\centering
    %{-1ex}
    \includegraphics[width=0.99\linewidth]{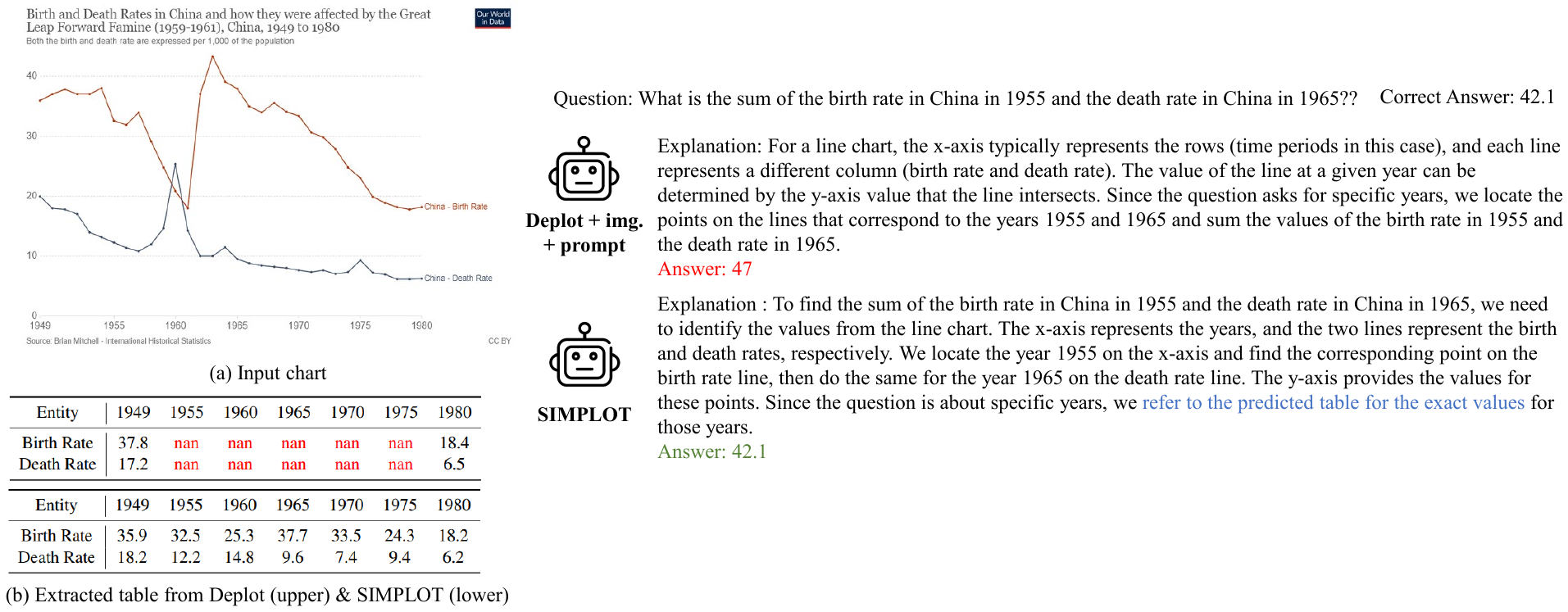} 
    %{-3ex}
    \caption{Case study. Effectiveness of precise extracted table for challenging question.}
    \label{fig:hardexample}
    %{-2.5ex}
\end{figure*}

Hence, we generated more challenging questions, and evaluated models in Section \ref{sec:furtheranalysis}. 
Fig. \ref{fig:hardexample} shows an example of challenging question, where Deplot fails to answer correctly due to its poorly extracted table, while \proposed~generates a precise answer based on its accurately extracted table. The question is \textit{"What is the sum of the birth rate in China in 1955 and the death rate in China in 1965?"}, which requires more than one row and column to answer. In this example, we verify that extracting accurate tables is crucial for solving challenging questions, and that the ChartQA dataset is insufficient to precisely evaluate the reasoning models.

\section{Error Analysis}\label{app:error}

\begin{figure*}[t]  %%% t: top, b: bottom, h: here
\centering
    %{-1ex}
    \includegraphics[width=0.99\linewidth]{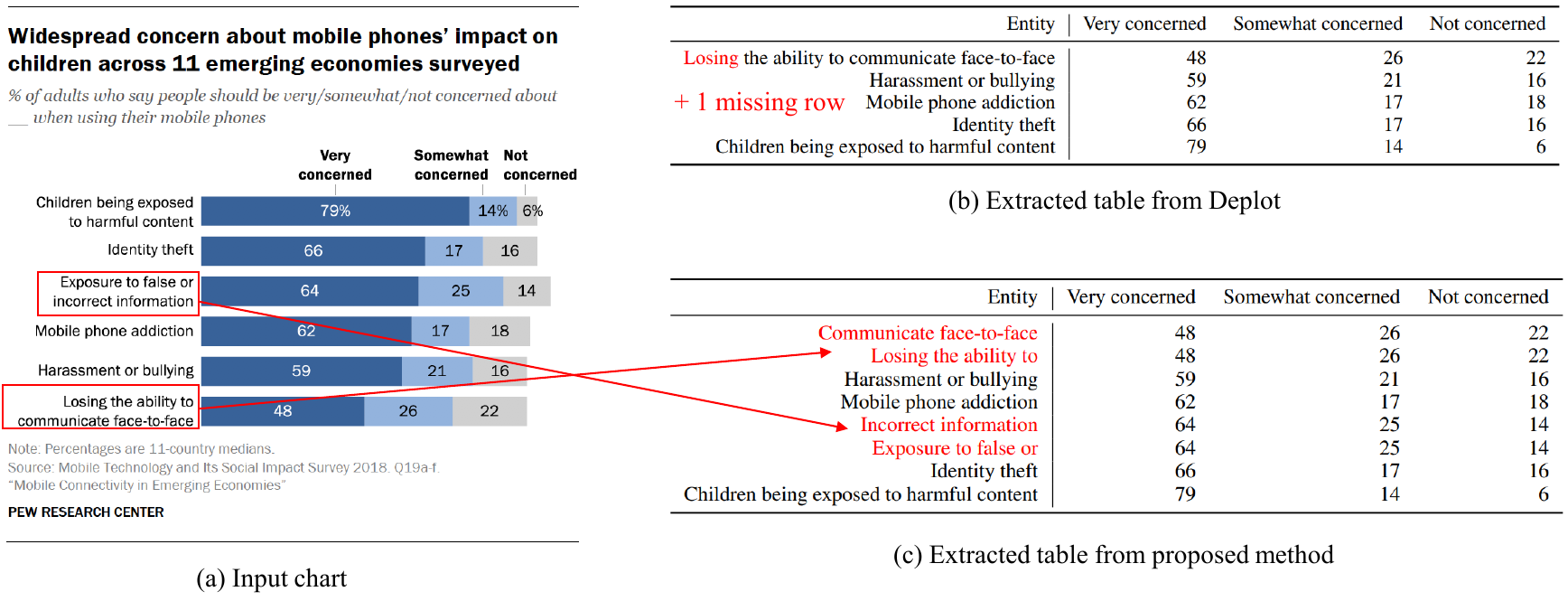} 
    %{-1ex}
    \caption{Error Analysis. Failure in row / column rendering.}
    \label{fig:error1}
    %{-2ex}
\end{figure*}

Here we present examples of \proposed~failures, their reasons, and possible solutions. In Fig. \ref{fig:error1}~(c), it can be observed that the extracted table from the proposed method contains inaccurate rows. This is attributed to  a line break between the items "Exposure to false or incorrect information" and "Losing the ability to Communicate face-to-face" in the input chart, leading to the misinterpretation of each item as two separate entities during row extraction. To address such issues, one possible solution would be to refine the rows and columns by considering their contextual information before rendering.

\begin{figure*}[!h]  %%% t: top, b: bottom, h: here
\centering
    %{-1ex}
    \includegraphics[width=0.99\linewidth]{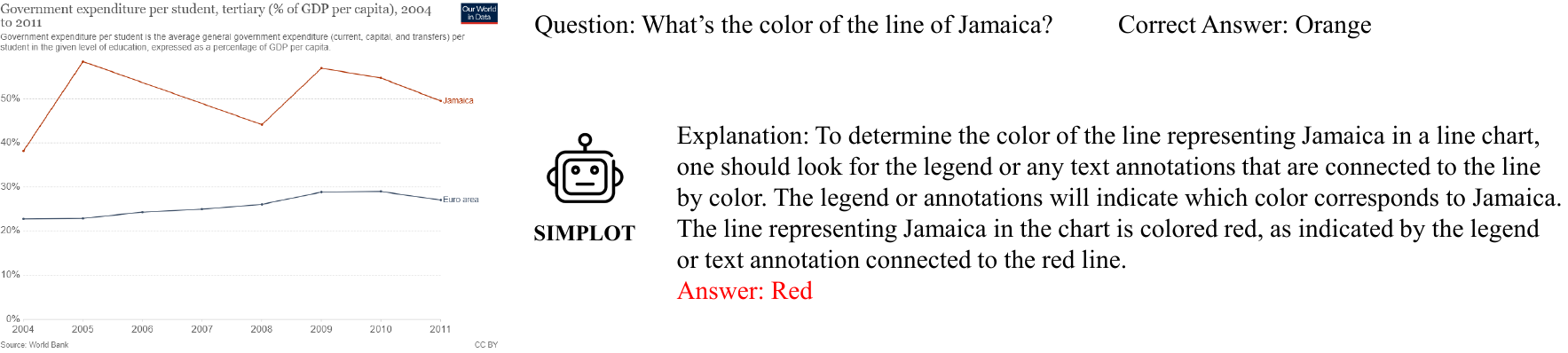} 
    %{-1ex}
    \caption{Error Analysis. Failure in detecting a color in the chart.}
    \label{fig:error2}
    %{-2ex}
\end{figure*}

In addition, we find some examples where the LMM lacks specific abilities, such as color detection. In Fig. \ref{fig:error2}, there is a question asking about the color of "Jamaica" with the correct answer being orange, but the model recognizes it as red and answers incorrectly. One possible solution would be to broaden the range of similar colors for the model to answer together.

\section{Responsible NLP Research Checklist}\label{app:checklist}

\subsubsection*{Hardware.}
We train \proposed~using a NVIDIA RTX A6000. The training time for phase 1 (Section \ref{sec:phase1}) and phase 2 (Section \ref{sec:phase2}) are about 2 hours and 4 hours per epoch, respectively. We train phase 1 for 7 epochs and phase 2 for 9 epochs.

\subsubsection*{Parameters.}
\proposed~has 374M parameters, and the exact number of parameters in GPT-4 has not been publicly reported. For LMM prompting, we use temperature of 0.1.

\begin{figure*}[!t]  %%% t: top, b: bottom, h: here
\centering
    %{-1ex}
    \includegraphics[width=1\linewidth]{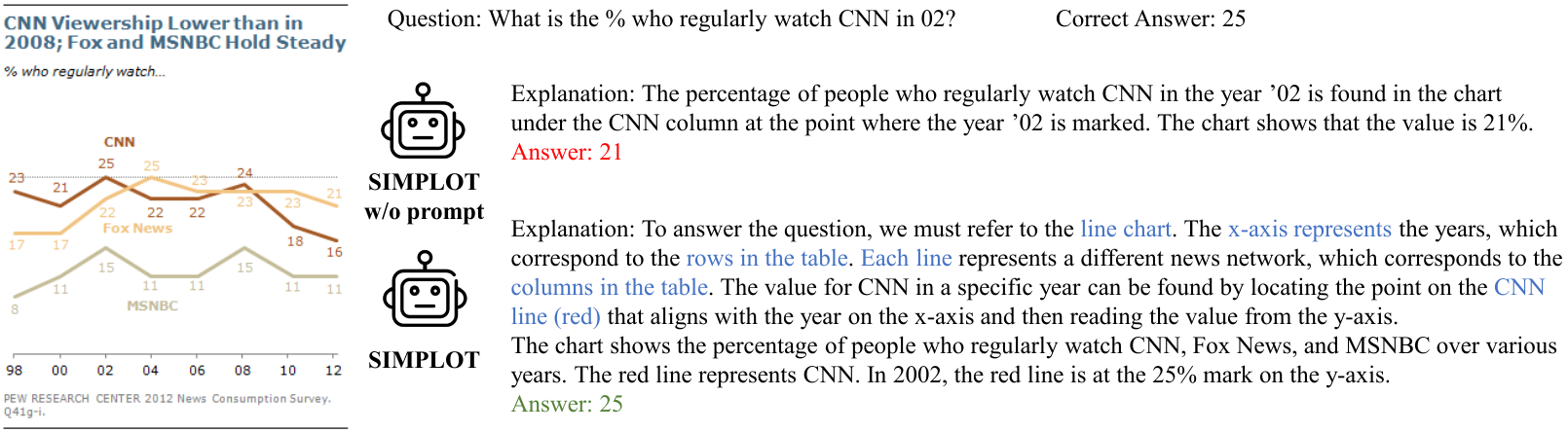} 
    %{-2ex}
    \caption{Case study. Interpreting charts similar to how humans do using \prompt.}
    \label{fig:case5}
    %{-2ex}
\end{figure*}

\end{document}